\documentclass{article}

\usepackage[numbers,square,sort&compress]{natbib}

\usepackage[preprint]{neurips_2026}

\usepackage[utf8]{inputenc} % allow utf-8 input
\usepackage[T1]{fontenc}    % use 8-bit T1 fonts
\usepackage{url}            % simple URL typesetting
\usepackage{booktabs}       % professional-quality tables
\usepackage{amsfonts}       % blackboard math symbols
\usepackage{nicefrac}       % compact symbols for 1/2, etc.
\usepackage{microtype}      % microtypography
\usepackage{xcolor}         % colors

\usepackage{amsmath,amsfonts,bm}

\def\eqref#1{equation~\ref{#1}}
\def\1{\bm{1}}

\def\rp{{\textnormal{p}}}

\def\rs{{\textnormal{s}}}

\def\ry{{\textnormal{y}}}

\def\rvp{{\mathbf{p}}}

\def\rvs{{\mathbf{s}}}

\def\rvy{{\mathbf{y}}}

\def\vtheta{{\bm{\theta}}}

\def\vx{{\bm{x}}}

\DeclareMathAlphabet{\mathsfit}{\encodingdefault}{\sfdefault}{m}{sl}
\SetMathAlphabet{\mathsfit}{bold}{\encodingdefault}{\sfdefault}{bx}{n}

\def\gA{{\mathcal{A}}}

\def\gD{{\mathcal{D}}}

\def\gP{{\mathcal{P}}}

\def\gS{{\mathcal{S}}}
\def\gT{{\mathcal{T}}}

\newcommand{\sigmoid}{\sigma}

\DeclareMathOperator*{\argmax}{arg\,max}

\usepackage{microtype}
\usepackage{graphicx,wrapfig}
\usepackage{subcaption}
\usepackage[acronym,toc,shortcuts]{glossaries}
\usepackage{xurl}
\usepackage[hidelinks]{hyperref}
\usepackage{url} 
\usepackage{lipsum}
\usepackage{siunitx}
\usepackage{multirow}
\usepackage{wrapfig}
\usepackage{caption}
\usepackage{comment}
\usepackage{svg}
\usepackage{array,ragged2e}
\newcolumntype{P}[1]{>{\RaggedRight\arraybackslash}p{#1}}

\usepackage{amsmath}
\usepackage{amssymb}
\usepackage{mathtools}
\usepackage{amsthm}
\usepackage[capitalize,noabbrev]{cleveref}

\newcommand{\entropy}{\mathcal{H}}
\newcommand{\KLdiv}{D_{\mathrm{KL}}}
\newcommand{\indicator}{\mathbb{I}}

\newcommand{\ie}{\textit{, i.e., }}
\newcommand{\eg}{\textit{, e.g., }}

\newcommand{\SubItem}[1]{
    {\setlength\itemindent{15pt} \item[-] #1}
}

\newacronym{ddpm}{DDPM}{Denoising Diffusion Probabilistic Model}
\newacronym{dit}{DiT}{Diffusion Transformer}
\newacronym{cama}{CAMA}{Cohort-based Active Modality Acquisition}
\newacronym{auprc}{AUPRC}{Area Under the Precision-Recall Curve}
\newacronym{auroc}{AUROC}{Area Under the Receiver Operating Characteristic}
\newacronym{sota}{SOTA}{state-of-the-art}
\newacronym{al}{AL}{Active Learning}
\newacronym{afa}{AFA}{Active Feature Acquisition}
\newacronym{ama}{AMA}{Active Modality Acquisition}
\newacronym{kldiv}{KL-Divergence}{Kullback-Leibler Divergence}
\newacronym{mlp}{MLP}{Multi-Layer Perceptron}
\newacronym{vit}{ViT}{Vision Transformer}
\newacronym{vae}{VAE}{Variational Auto Encoder}
\newacronym{ae}{AE}{Auto Encoder}
\newacronym{gan}{GAN}{Generative Adversarial Network}
\newacronym[
  shortplural={SEMs},
  longplural={standard errors of the mean}
]{sem}{SEM}{standard error of the mean}
\newacronym{mse}{MSE}{mean squared error}
\newacronym{rl}{RL}{Reinforcement Learning}
\newacronym{llm}{LLM}{Large Language Model}
\newacronym{mri}{MRI}{magnetic resonance image}
\newacronym{ecg}{ECG}{electrocardiogram}
\newacronym{af}{AF}{acquisition function}
\newacronym{ukbb}{UKB}{UK Biobank}
\newacronym{bcvae}{BC-VAE}{Beta-Conditional-VAE}
\newacronym{sle}{SLE}{systemic lupus erythematosus}
\newacronym{ehr}{EHR}{electronic health record}
\newacronym{prs}{PRS}{polygenic risk scores}
\newacronym{bnp}{BNP}{B-type natriuretic peptide}
\newacronym{lf}{LF}{late fusion}
\newacronym{fid}{FID}{Fréchet inception distance}
\newacronym{eig}{EIG}{expected information gain}
\newacronym{elbo}{ELBO}{evidence lower bound}

\title{Cohort-Based Active Modality Acquisition}

\author{
  \textbf{Tillmann Rheude\textsuperscript{1,3$^\dagger$}}, 
  \textbf{Roland Eils\textsuperscript{1,2,3$^\dagger$}}, 
  \textbf{Benjamin Wild\textsuperscript{1$^\dagger$}}\\
  \textsuperscript{1}Berlin Institute of Health, Charité - Universitätsmedizin Berlin, 
  \textsuperscript{2}Intelligent Medicine Institute,\\Fudan University, 
  \textsuperscript{3}Department of Mathematics and Computer Science, Freie Universität Berlin\vspace{2mm} \\
  \textsuperscript{$^\dagger$}\texttt{\{benjamin.wild, roland.eils, tillmann.rheude\}@bih-charite.de}
}

\begin{document}

\maketitle

\begin{abstract}
    %Real-world machine learning applications often involve data from multiple modalities that must be integrated effectively to make robust predictions. However, in many practical settings, modalities can be missing and their acquisition can be costly.
    %This raises the question: which samples should be prioritized for modality acquisition given limited resources?
    Real-world multimodal machine learning often faces missing, costly-to-acquire modalities, raising the problem of which samples to prioritize for additional acquisition under a budget.
    % While prior work has explored individual-level acquisition strategies and training-time active learning paradigms, test-time and cohort-based acquisition remain underexplored.
    Prior work mainly studies per-sample or training-time acquisition while test-time, cohort-level acquisition is less explored.
    %We introduce \ac{cama}, a novel test-time setting to formalize the challenge of selecting which samples should receive an additional modality.
    %We derive acquisition strategies that leverage a combination of generative imputation and discriminative modeling to estimate the expected benefit of acquiring a missing modality based on common evaluation metrics.
    %We also introduce upper-bound heuristics that provide performance ceilings to benchmark acquisition strategies. 
    We propose \ac{cama}, a novel test-time cohort-level modality acquisition setting, and introduce imputation-based acquisition strategies that estimate the expected utility of acquiring a missing modality, along with upper-bound heuristics for benchmarking.
    Experiments on datasets with up to 15 modalities demonstrate that our proposed imputation-based strategies can more effectively guide the acquisition of an additional modality for selected samples compared with methods relying solely on pre-acquisition information, entropy-based guidance, or random selection. 
    We showcase the real-world relevance and scalability of our method by demonstrating its ability to guide the acquisition of proteomics data for disease prediction in a large prospective cohort, the \ac{ukbb}. Our work provides an effective approach for optimizing modality acquisition at the cohort level, enabling more effective use of resources in constrained settings.\footnote{Code will be published on GitHub.}
\end{abstract}

\section{Introduction}
\label{sec:introduction}

Consider a clinical healthcare setting where all patients in a cohort undergo a standard, inexpensive set of initial examinations, such as basic blood tests and anamnesis. However, a more advanced, expensive, or invasive procedure, like genomic sequencing or specialized imaging, could offer crucial diagnostic or prognostic information for a subset of these patients \citep{huang_what_2021}. Given a limited budget or capacity for the more advanced procedure, the central question becomes: which patients should receive this additional resource to maximize the overall diagnostic yield or improve treatment outcomes across the entire cohort? 
\begin{figure*}[t]
    \centering
    \includegraphics[width=.99\linewidth]{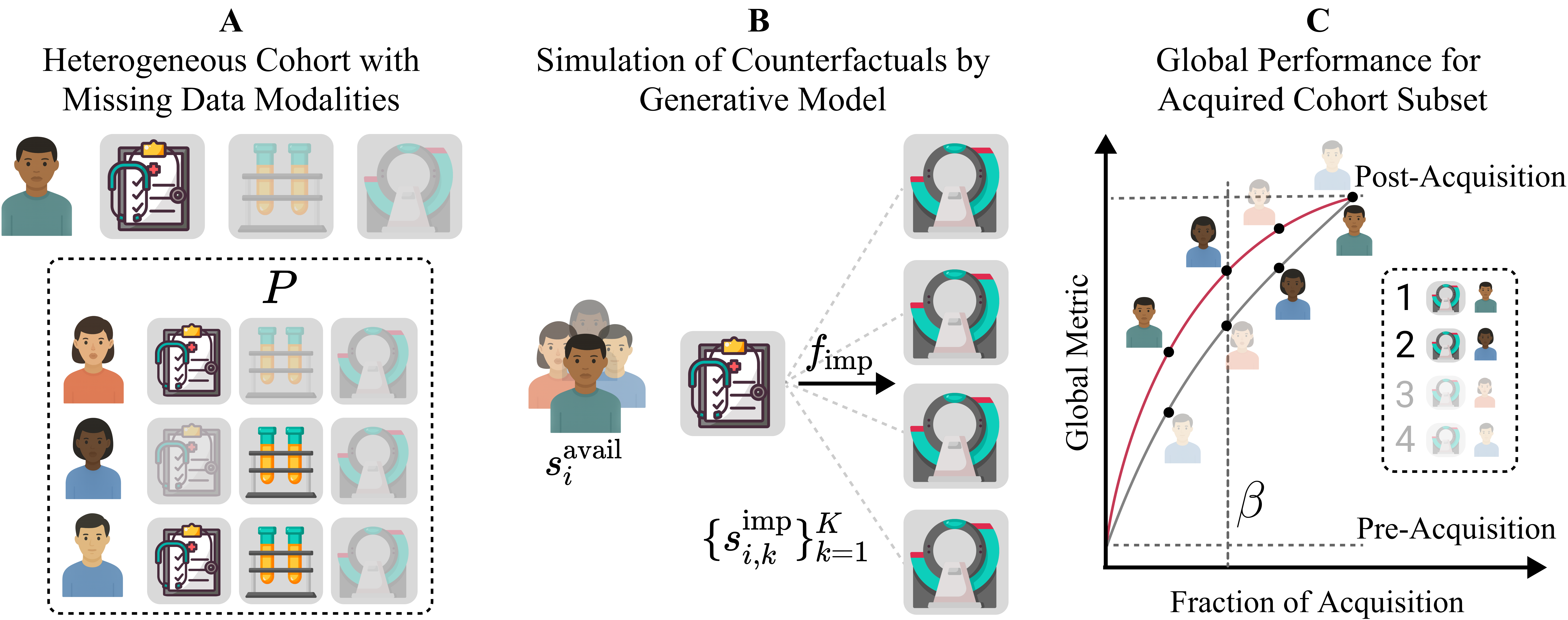}
    \caption{Motivational example for \ac{cama} determining the added value of obtaining the \ac{mri} modality.
    \textbf{(A)} A heterogeneous cohort for which each sample has  $P$ distinct modalities.
    \textbf{(B)} Instead of using the initial subset logit scores $\rs_i^{\textnormal{avail}}$, a generative model $f_{\textnormal{imp}}$ imputes the target missing modality for every patient in the cohort. This yields imputed, augmented-modality logit scores $\{\rs_{i,k}^{\textnormal{imp}}\}_{k=1}^K$ that approximate the logits as if that modality were available. These scores approximate $\rs_i^{\textnormal{acquired}}$\ie the counterfactual with only the imputed modality added.
    \textbf{(C)} An \ac{af} utilizes these scores to rank samples by acquisition priority. The graph demonstrates how the global performance metric improves from the initial baseline towards the performance of a model with access to post-acquisition data, as an increasing fraction of the cohort receives the additional modality. This acquisition process is guided by the proposed strategies operating under the acquisition budget constraint $\beta$.}
    \label{fig:illustrative_example}
\end{figure*}

For healthcare, budgets are often resource-specific rather than flexible. For example, a hospital may have a fixed capacity for one \ac{mri} scanner, or a cohort may have a specific grant for one modality. The critical decision is prioritizing access to that single resource across the cohort, and not necessarily dynamically acquiring for different modalities per patient. Concretely, for the UKB, only $10\%$ of the cohort do contain the proteomics modality. Notably, the proteomics modality was financed through a multi-million dollar pre-competitive consortium of $13$ pharmaceutical companies, and a $\pounds 20$ million government top-up was later needed to finish the full cohort acquisition \cite{Proteomics_20m,Vanderbilt_Biomarker_Core_Rate,Proteomics_Landmark_Step}. This highlights the remarkable need of \ac{cama} in real-life applications\ie the acquisition of a single, fixed modality for multimodal learning setups. 
Consider a healthcare system that can afford 1,000 expensive tests for a 100,000-person cohort. The goal is to improve health outcomes across the whole population, and this typically happens through resource allocation: who receives preventive interventions, who gets enrolled in clinical trials, who is flagged for closer monitoring. These decisions depend on accurate risk stratification. A global ranking of all 100,000 individuals by predicted risk becomes the tool through which such allocation decisions are made. The question is: which 1,000 patients should we test so that our final ranking of all 100,000 is as accurate as possible?

Balancing potential gains from data modalities against the costs and complexities of acquisition is not unique to healthcare. In remote sensing, for instance, decisions must be made regarding which geographical areas warrant costly high-resolution satellite imagery to supplement widely available, lower-resolution data. Likewise, in industrial quality assurance, manufacturers could decide which components from a production batch should undergo detailed, time-consuming testing in addition to rapid, standard visual inspections to effectively identify defects at a batch level.
The topic of efficient data acquisition has led to several established paradigms in machine learning, such as \mbox{\ac{al}} \mbox{\citep{holzmuller_framework_2023}}, \ac{afa} \citep{shim_joint_2018}, \ac{ama} \citep{kossen_active_2023}, and multimodal learning with missing data \citep{wu_deep_2024}. However, previous research predominantly centers on optimizing acquisition for individual samples and often does not directly address test-time budget constraints for an entire cohort.
Consequently, the strategic, test-time acquisition of an additional modality from a cohort perspective remains a significant, largely unaddressed gap. This setting involves deciding, for a given batch of new samples where different subsets of modalities are available, which specific samples should receive an additional, costly modality to best achieve a global objective\eg maximizing overall predictive performance or diagnostic accuracy for the cohort, subject to budget constraints. We hypothesize that imputation-based \acfp{af} can effectively guide resource allocation under cohort-level constraints. The main contributions of this work are as follows:

\begin{itemize}

    \item \textbf{The \ac{cama} setting} We introduce and formalize \ac{cama}, a previously unexplored setting that addresses the challenge of prioritizing which samples within a test-time cohort should undergo additional modality acquisition based on an available subset of modalities.

    \item \textbf{Development of \acp{af} for \ac{cama}} We propose a theoretical framework, derived from established evaluation metrics\eg \ac{auroc} and \ac{auprc}, that provides a foundation for developing \acp{af} within the \ac{cama} setting.

    \item \textbf{Architectures for \ac{cama}} We develop a novel architecture for approaching \ac{cama}, including a) derivations of \acp{af} by combining generative and discriminative deep learning and b) the definition of corresponding upper bounds to serve as performance benchmarks.

    \item \textbf{Comprehensive evaluation} We present a comprehensive empirical evaluation of our proposed methods across several multimodal datasets, which vary in their number of modalities and application domains, with up to 100,000 samples and 15 modalities. This includes an analysis of key assumptions, upper bounds and oracle strategies, performance challenges, and robustness.

\end{itemize}

\section{Related Work}
\label{sec:related_work}
In the following, we contextualize our work on \ac{cama} by reviewing the key concepts and contributions from several relevant research domains summarized briefly in \Cref{tab:related_work_summary}. 

\begin{table*}[h]
    \centering
    \small 
    \caption{Comparison of acquisition paradigms. \ac{cama} is unique due to cohort-level acquisition.}
    \label{tab:related_work_summary}
    
    \begin{tabular}{@{}lcccc@{}}
    \toprule
    \textbf{Paradigm} & \textbf{Acquisition} & \textbf{Decision Level} & \textbf{Time} & \textbf{Primary Objective} \\
    \midrule
    \acs{al} & Labels & Individual & Training & Maximize model performance \\
    \acs{afa} & Features & Individual & Test & Optimize sample-level prediction \\
    \acs{ama} & Modalities & Individual & Test & Optimize sample-level prediction \\
    \midrule
    %\textbf{CAMA (Ours)} & \textbf{Modalities} & \textbf{Cohort} & \textbf{Test} & \textbf{Maximize global cohort metric} \\
    \acs{cama}  & Modalities & Cohort & Test & Maximize global cohort metric \\
    \bottomrule
    \end{tabular}
\end{table*}

\paragraph{\acf{al}}
\ac{al} seeks to enhance model training by selecting unlabeled data points for annotation by an oracle \citep{settles_active_2012,ren_survey_2022,li_survey_2025}. Our methodology draws significantly from \ac{al} principles, particularly in the development of an \ac{af} to guide the selection process. Consequently, established \ac{al} strategies and concepts, such as those rooted in measuring uncertainty \citep{settles_active_2012,han_active_2021,hoarau_reducing_2025,raj_convergence_2022,ma_eddi_2019} or using generative models \citep{tran_bayesian_2019,zhu_generative_2017,zhang_generative_2024,ma_eddi_2019,peis_missing_2022}, are central to our work. Existing work on multimodal acquisition \citep{rudovic_multi-modal_2019,das_mavic_2022}, batch-level selection \citep{ash_deep_2020,kirsch_batchbald_2019,holzmuller_framework_2023}, and balanced \ac{al} \citep{aggarwal_active_2020,shen_towards_2023,zhang_algorithm_2023,hoarau_reducing_2025} is especially relevant. Our approach diverges from the conventional goals of directly optimizing model training or seeking labels for specific data points: We aim to identify samples for which the acquisition of an additional data modality would be most beneficial given cohort-level constraints.

\paragraph{\acf{afa}}
\ac{afa} builds upon \ac{al} by selecting the most informative individual features for a given sample, often considering their acquisition costs \citep{rahbar_survey_2025}. Similar to \ac{al} approaches, methods for \ac{afa} encompass a diverse range of techniques, including strategies based on measuring uncertainty \citep{hoarau_reducing_2025,astorga_active_2024}, the use of generative models \citep{li_active_2021,li_distribution_2024,gong_icebreaker_2019,zannone_odin_2019}, and \ac{rl} \citep{valancius_acquisition_2024,janisch_classification_2020,kleist_evaluation_2025,shim_joint_2018,baja_measure_2025}. Other common methodologies involve batch-level perspectives \citep{asgaonkar_generator_2024}, leveraging information bottlenecks \citep{norcliffe_information_2025}, or employing the \ac{kldiv} \citep{natarajan_whom_2018}. Some \ac{afa} techniques rely on gradient calculations \citep{ghosh_difa_2023}, while distinct approaches are formulated as individual, sequential recommender systems \citep{freyberg_mint_2024,martel_peri-diagnostic_2020}. At an application level, even \acp{llm}, such as Med-PaLM~2 \citep{singhal_toward_2025}, could be employed for \ac{afa}, although such deployments remain unexplored in this context. While our setting shares the core idea of \ac{afa}, it differs significantly: We are not concerned with the selection of individual features, but rather with identifying which entire data modalities to acquire. Furthermore, this decision-making process is applied at the cohort level, rather than optimizing for individual samples.

\paragraph{\acf{ama}}
\ac{ama} can be conceptualized as an extension of \ac{afa}, distinguished by its focus on selecting entire data modalities rather than individual features or labels. Prominent related research includes approaches employing \ac{rl} for multimodal data \citep{kossen_active_2023,jain_test-time_2025,li_towards_2025} and methods utilizing submodular optimization in conjunction with Shapley values \citep{kuhn_value_1953,he_efficient_2024}. The approach by \citet{kossen_active_2023} differs from ours through its reliance on \ac{rl}, whereas \citet{he_efficient_2024} primarily investigate how modalities affect optimal learning performance. Further studies have explored the use of Gaussian mixtures within Bayesian optimal experimental design to enhance data acquisition efficiency for model training \citep{long_multimodal_2022}. This objective differs from ours, as our focus is not on improving the model training process itself, but rather on optimizing performance for a downstream task at test time. The relative sparsity of existing work for \ac{ama} underscores the significance of the research gap that our proposed setting\ie \ac{cama}, aims to address.

\paragraph{Multimodal Learning with Missing Data Modalities}
Research in multimodal learning with missing data modalities offers techniques for robustly handling incomplete datasets. These methods are broadly classified into strategy design aspects\ie architecture-focused designs and model combinations, and data processing aspects\ie representation learning and modality imputation \citep{wu_deep_2024}. Acknowledging the utility of these approaches, our work emphasizes imputation-based strategies, and thus this paragraph highlights those methods. Imputation of missing features is commonly performed using \acp{ae} \citep{hinton_autoencoders_1993}, \acp{vae} \citep{kingma_auto-encoding_2014}, \acp{gan} \citep{goodfellow_generative_2014}, or \acp{ddpm} \citep{ho_denoising_2020,rombach_high-resolution_2022}. These methods naturally extend to multiple modalities, for example, with \ac{vae}-based \citep{wesego_score-based_2024,sutter_generalized_2021,lewis_accurate_2021} and \ac{ddpm}-based \citep{wang_incomplete_2023} approaches. Notably, the latter\ie \textit{IMDer} \citep{wang_incomplete_2023}, a multimodal deep learning architecture that imputes missing values with \acp{ddpm} in latent spaces, is adapted in our work (\Cref{sec:evaluation}). However, this research area focuses on handling absent modalities rather than deciding which ones to acquire.

\section{Problem Formulation}
\label{sec:problem_formulation}

Let $\gD = \{ (\vx_i, \ry_i) \}_{i=1}^{N}$ be a dataset of $N$ samples. For each sample $i$, the full feature set $\vx_i$ is composed of $P$ distinct data modalities, $\vx_i = \{\vx_i^{(1)}, \dots, \vx_i^{(P)}\}$, and $\ry_i \in \{0, 1\}$ is the corresponding binary label. In practice, only a subset of these modalities may be available. We denote the set of indices of available modalities for sample $i$ as $\gP_i^{\textnormal{avail}} \subseteq \{1, \dots, P\}$. Our goal is to decide for which samples to acquire costly missing data to maximize a cohort-level performance metric. This decision is guided by predictive scores (logits), and we consider three key predictive scores for each sample $i$:

\begin{itemize}
    \item $\rs_i^{\textnormal{avail}}$: The available score, computed using the subset of data modalities that are already observed.
    % \item $\rs_i^{\textnormal{full}}$: The full score, a theoretical oracle score that would be computed if all modalities were available. This is unknown for samples with missing data modalities.
    \item $\rs^{\text{acquired}}_{i}$: The acquired score, computed using the available modalities and the newly acquired modality.
    \item $\{\rs_{i,k}^{\textnormal{imp}}\}_{k=1}^K$: A set of $K$ imputed scores that estimate the unknown $\rs_i^{\textnormal{acquired}}$ using the available data modalities.
\end{itemize}

For instance, given the example from \Cref{fig:illustrative_example}, in a simple clinical setting with a cheap, universally available base modality\eg cardiac biomarkers such as troponin or \ac{bnp}, and an expensive additional modality\eg cardiac \ac{mri}, $\rs_i^{\textnormal{avail}}$ would be the score from the blood tests alone, while $\rs_i^{\textnormal{acquired}}$ would be the score using both tests and \ac{mri}. To compute these scores, we assume a single model $f$ parameterized by $\vtheta$ that can process any subset of modalities. The available and acquired scores are thus:
\begin{align}
    \rs_i^{\textnormal{avail}} &= f(\vx_i^{\textnormal{avail}}, \vtheta) \\
    \rs_i^{\textnormal{acquired}} &= f(\vx_i^{\textnormal{acquired}}, \vtheta)
\end{align}
where $\vx_i^{\textnormal{avail}}$ and $\vx_i^{\textnormal{acquired}}$ represent the feature sets for the available and acquired modalities, respectively. To estimate the acquired score without costly acquisition, we use a generative imputation model $f_{\textnormal{imp}}$. This model generates a set of $K$ plausible embeddings that enable the classifier $f_C$ to predict the scores $\{\rs_{i,k}^{\textnormal{imp}}\}_{k=1}^K$. These imputation-based scores form the basis of our acquisition functions.

%\subsection{Acquisition and Optimization}
%\label{subsec:acquisition_and_optimization}
The goal of the optimization is to select a subset of samples $\gS$ from the cohort of $N$ total samples for which an additional modality should be acquired. This subset $\gS \subseteq \{1, \dots, N\}$ has a predetermined size $|\gS| = \beta$, where $\beta$ is the acquisition budget\ie the number of samples for which additional modalities will be acquired. The final score $\rs_i(\gS)$ used for the evaluation of a sample $i$ is then determined by the selection:
\begin{align}
    \rs_i(\gS) = \begin{cases} \rs_{i}^{\textnormal{acquired}} & \text{if } i \in \gS \\ \rs_{i}^{\textnormal{avail}} & \text{if } i \notin \gS \end{cases}.
\end{align}

The optimization problem is to find the set $\gS^{*}$ that maximizes the chosen performance metric:
\begin{align}
    \gS^* = \argmax_{\gS \subseteq \{1, \dots, N\} : |\gS|=\beta} \text{Metric}(\rvy, \rvs(\gS))
    \label{eq:optimization_performance_metric}
\end{align}
where $\rvy=\{\ry_i\}_{i=1}^N$ is the vector of true labels, and $\rvs(\gS) = \{\rs_i(\gS)\}_{i=1}^N$ is the vector of resulting scores for all samples in the cohort. The task is to identify an optimal, constrained subset for which to acquire additional modalities, while maximizing a performance metric across the entire cohort. 

\section{Acquisition Function Strategies}
\label{sec:optimizing_metrics_and_acquisition_functions}
\begin{table}[t]
    \centering
    \small
    \caption{
        Summary of \acf{af} strategies.
    }
    \label{tab:af_summary}
    \begin{tabular}{@{} P{1.3cm} P{2.9cm} P{3.8cm} P{4.7cm} @{}}
        \toprule
        \textbf{Category} & \textbf{Strategies} & \textbf{Inputs} & \textbf{Ranking Criteria} \\
        \midrule
        
        Oracle & 
        \ac{auroc}, \ac{auprc} & 
        True labels \& acquired scores & 
        Greedy selection for maximum gain. \\
        \addlinespace

        Upper-Bound & KL-Div., Rank, Uncertainty & True acquired scores & True change in prediction, cohort rank, or uncertainty. \\
        \addlinespace

        Imputation & KL-Div., Rank, Uncertainty \& Probability & Imputed acquired scores & Expected change in prediction, rank, or uncertainty. \\
        \addlinespace
        
        Baselines & Uncertainty, Probability & Pre-acquisition scores & Uncertainty or probability using the available modality. \\
        \addlinespace

        Random & Random & None & Random selection. \\

        \bottomrule
    \end{tabular}
\end{table}
Directly solving the optimization problem to identify the optimal sample set $\gS^*$ is computationally intractable. Therefore, we employ heuristic \acfp{af} that approximate the optimal selection by ranking samples for modality acquisition. These strategies (\Cref{app:details_about_acquisition_function_strategies}) are derived from standard discriminative metrics (\Cref{subsec:auroc_and_auprc}) and can be categorized as follows (\Cref{tab:af_summary}):

\begin{itemize}
    \item \textbf{Oracle Strategies:} As upper-bound benchmarks, they assume perfect knowledge of outcomes and true labels to greedily select samples yielding the largest gain in the metric.
    
    \item \textbf{Upper-Bound Heuristic Strategies:} These heuristics assume knowledge of scores under modality completion but are label-agnostic, relying on metrics like the true uncertainty reduction, rank change, or \ac{kldiv}.
    
    \item \textbf{Imputation-Based Strategies:} Grounded in counterfactual reasoning, these strategies use a generative model to predict how the score might change if a missing modality were acquired.
    
    \item \textbf{Baseline Information Strategies:} These strategies only use information from the available modalities\ie without imputations, such as its predicted uncertainty or probability.
    
    \item \textbf{Random Strategy:} A baseline by selecting samples randomly, without any model scores.
\end{itemize}

Intuitively, these \acp{af} approximate the \ac{eig} from acquiring an additional modality. In our setting, \ac{eig} quantifies the expected improvement in a chosen performance metric given the additional information that would become available through a new modality. 
%\paragraph{Normalized Area of Gain}
For evaluation, we introduce a metric that describes the cumulative performance of an \ac{af}, normalized by the total possible gain achievable by transitioning all samples to post-acquisition performance (see \Cref{fig:illustrative_example} C, for an illustrative curve). Let $M_{\text{AF}}(b)$ denote the performance curve of an \ac{af} strategy for a primary metric $M$ as a function of the budget fraction $b$ of the acquisition budget $\beta$, $M_{\text{pre}}$ the performance of the pre-acquisition baseline, and $M_{\text{post}}$ the performance of the post-acquisition model. 
The normalized area of gain for an \acl{af} $\text{\ac{af}}$, which measures the portion of achievable performance gain captured across different budgets, is defined as the area under the performance gain curve, normalized by the maximum possible gain (\Cref{eq:g_full}). Intuitively, a value of $0$ indicates no improvement over the pre-acquisition baseline across budgets. A value of $1$ indicates matching the post-acquisition performance on average across budgets. Values greater than $1$ occur when the cohort’s performance at intermediate budgets temporarily exceeds the post-acquisition cohort as detailed later and shown in \Cref{fig:auroc_curves_side_by_side}.

\begin{align}
    G_{\text{full}}^{M}(\text{AF}) = \frac{\int_{0}^{1} (M_{\text{AF}}(b) - M_{\text{pre}}) \, db}{M_{\text{post}} - M_{\text{pre}}}
    \label{eq:g_full}
\end{align}

%\clearpage 
%\newpage 
\section{Experiments}
\label{sec:evaluation}
\begin{figure*}[t]
    \centering
    \includegraphics[width=.99\linewidth]{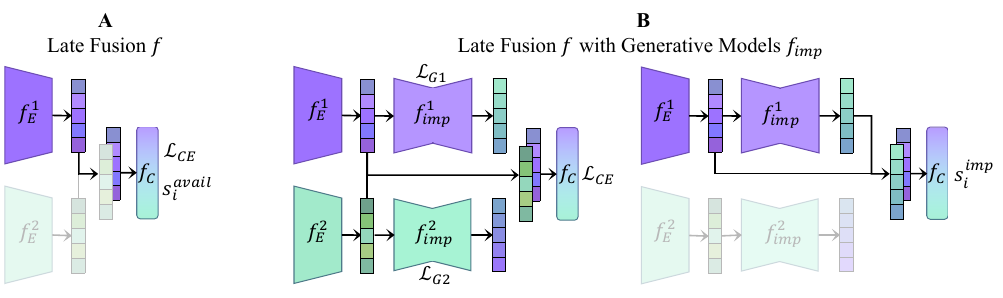}
    \caption{End-to-end architectures to determine the scores for different \acp{af} in our proposed \ac{cama} setting.
    \textbf{(A)} Vanilla \acf{lf} architecture of a model $f$ that can handle missing data modalities by masking. The model creates scores $s_i^{\text{avail}}$ given the available modalities. 
    %If all modalities are available, the model creates scores $s_i^{\text{full}}$, otherwise scores $s_i^{\text{avail}}$ are created.
    \textbf{(B)} Architecture for training (left) and inference (right) with a \ac{lf} model $f$ and a generative model $f_{\text{imp}}$ to create scores $s_i^{\text{imp}}$ for the imputation-based \acp{af}.}
    \label{fig:model_architecture}
\end{figure*}
To evaluate the \acp{af}, we require architectures that produce the necessary scores. The oracle, upper-bound, baseline, and random \acp{af} can be evaluated using a vanilla discriminative \acf{lf} model (\Cref{fig:model_architecture} A), as they operate on true labels $\ry_i$ and true scores $\rs_i^{\textnormal{acquired}}$ and $\rs_i^{\textnormal{avail}}$ (\Cref{sec:problem_formulation}). In contrast, our proposed imputation-based \acp{af} require the model to predict how its output would change if a missing modality were present. This necessitates an architecture that combines the discriminative component with a generative component to impute the missing modality (\Cref{fig:model_architecture} B).

\paragraph{Model Architecture}
First, we implement an architecture with modality-specific encoders $f_E^{(m)}: \mathcal{X}^{(m)} \to \mathbb{R}^d$ and a fusion classifier $f_C$ (\Cref{fig:model_architecture} A). The encoders map raw inputs for a sample $i$ and modality $m$ to latent embeddings $z_i^{(m)} = f_E^{(m)}(x_i^{(m)})$\eg with a \ac{vit} \citep{dosovitskiy_image_2021} for images and BERT \citep{devlin_bert_2019} for text. The classifier $f_C$ aggregates these embeddings to produce logits $s_i^{\text{avail}}$ depending on the availability of the raw inputs\ie with a Transformer encoder \citep{vaswani_attention_2017}. 
Second, for the imputation-based \acp{af}, we incorporate generative modules additionally to the discriminative late fusion (\Cref{fig:model_architecture} B) \citep{wang_incomplete_2023}. The generative modules $f_{\text{imp}}$ are parameterized as \acp{dit} \citep{peebles_scalable_2023} or \acp{bcvae} \citep{higgins_beta-vae_2017} trading off performance vs. efficiency. They are trained to approximate the conditional distribution $p_\theta(z^{(k)} | \{z^{(m)}\}_{m \in \mathcal{P}^{\text{avail}}_i})$ for each target modality $k$. The generative loss for sample $i$ is:
\begin{equation}
    \mathcal{L}_{G_i} = \sum_{k \in \mathcal{P}^{\text{avail}}_i} \mathcal{L}_{\text{gen}}\!\left(z_i^{(k)};\, \{z_i^{(m)}\}_{m \in \mathcal{P}^{\text{avail}}_i}\right)
\end{equation}
where $\mathcal{L}_{\text{gen}}$ is a variational bound on the negative conditional log-likelihood: for \acp{ddpm}, this corresponds to the denoising objective \citep{ho_denoising_2020}; for \acp{bcvae}, this is the negative conditional \ac{elbo}
\citep{higgins_beta-vae_2017}. For the discriminative task, the loss for sample $i$ is defined as binary cross entropy loss with the label $y$ and the predicted probability $p$:
\begin{equation}
    \mathcal{L}_{\text{CE}_i} = - \left[ y_i \log(p_i) + (1 - y_i) \log(1 - p_i) \right].
\end{equation}
The final loss function for the whole architecture is defined as the combination of both loss terms:
\begin{align}
    \mathcal{L} = \lambda_1 \mathcal{L}_{\text{CE}} + \lambda_2 \mathcal{L}_{\text{G}}
\end{align}
with loss weightings $\lambda$. During inference, the classifier $f_C$ also uses samples of $p(z^{(k)} | \{z^{(m)}\}_{m \in \mathcal{P}^{\text{avail}}_i})$ for missing modalities to create the scores $s_i^{\text{imp}}$ needed for our generative \acp{af} (\Cref{fig:model_architecture} B, right). During training, samples from $p(z^{(k)} | \{z^{(m)}\}_{m \in \mathcal{P}^{\text{avail}}_i})$ are not passed to $f_C$, even when modalities are missing (\Cref{fig:model_architecture} B, left). Instead, the discriminative components ($f_E^{(m)}$ and $f_C$) are trained only on available modalities via attention masks. This means that the generative and discriminative parts are trained jointly, but the generative outputs do not directly influence the classifier during training beyond the shared encoders being updated by the classification loss. For model training, we use the ScheduleFree optimizer \citep{defazio_road_2024}.
We find the following architectural decisions essential (\Cref{tab:ablation}): (a) Layer Normalization \citep{ba_layer_2016} at the end of each modality's encoder to stabilize the generative models operating between latent spaces, (b) label smoothing \citep{szegedy_rethinking_2016} to produce less overconfident and better-calibrated probability distributions, (c) decoupling the generative modules from the classifier during training and (d) class balancing the training dataset (next paragraph).
Regarding missing modalities, we do not pre-train on all available data modalities, in contrast to \citet{wang_incomplete_2023}. We use a predefined, seed-dependent missing-modality mask to control data modality leakage during training unlike batch-dependent masks, which eventually reveal all modalities for every sample across numerous epochs. Details in \Cref{app:hyperparameter_sweeps}.

\paragraph{Datasets}
We evaluate \ac{cama} on the \ac{ukbb} \citep{sudlow_uk_2015}, MIMIC Symile \citep{saporta_contrasting_2024}, MIMIC HAIM \citep{soenksen_integrated_2022,soenksen_code_2022}, and MOSEI \citep{zadeh_multimodal_2018}, which cover diverse domains such as healthcare and emotion recognition.
Missing modalities are synthetically created, except for the \ac{ukbb}. We design the datasets for binary classification, resulting in ten binary targets for the MIMIC datasets and one binary target for MOSEI and \ac{ukbb}. 
%While MOSEI is already class-balanced \citep{zadeh_multimodal_2018}, HAIM and Symile exhibit significant class imbalance \citep{soenksen_integrated_2022,zadeh_multimodal_2018}. 
We employ random oversampling during training for the originally unbalanced HAIM and Symile \citep{soenksen_integrated_2022,zadeh_multimodal_2018}.
Importantly, during testing we retain the original imbalanced distributions.
We highlight \ac{ukbb} as the most challenging dataset to demonstrate that \ac{cama} scales to a broad multimodal range and large-scale cohorts with approximately $100{,}000$ samples and $15$ modalities. We focus on acquiring the costly proteomics modality for predicting the onset of \ac{sle}, which benefits from the proteomics acquisition \citep{yang_plasma_2025}. Details in \Cref{app:dataset_details}.

\begin{figure}[t]
    \centering

    \begin{subfigure}{0.48\textwidth}
        \includegraphics[width=\linewidth]{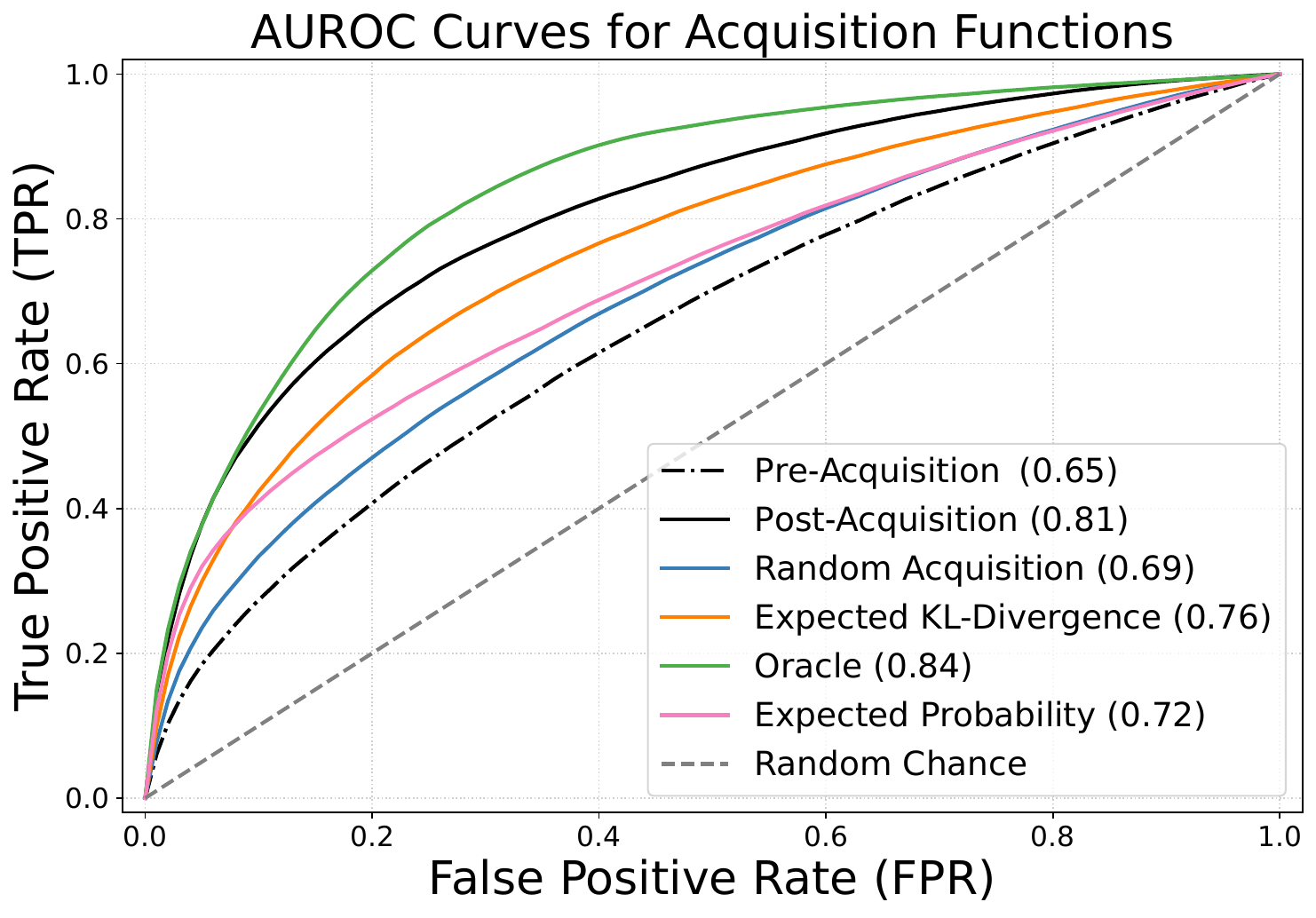}
        \caption{}
    \end{subfigure}
    \hfill
    \begin{subfigure}{0.48\textwidth}
        \includegraphics[width=\linewidth]{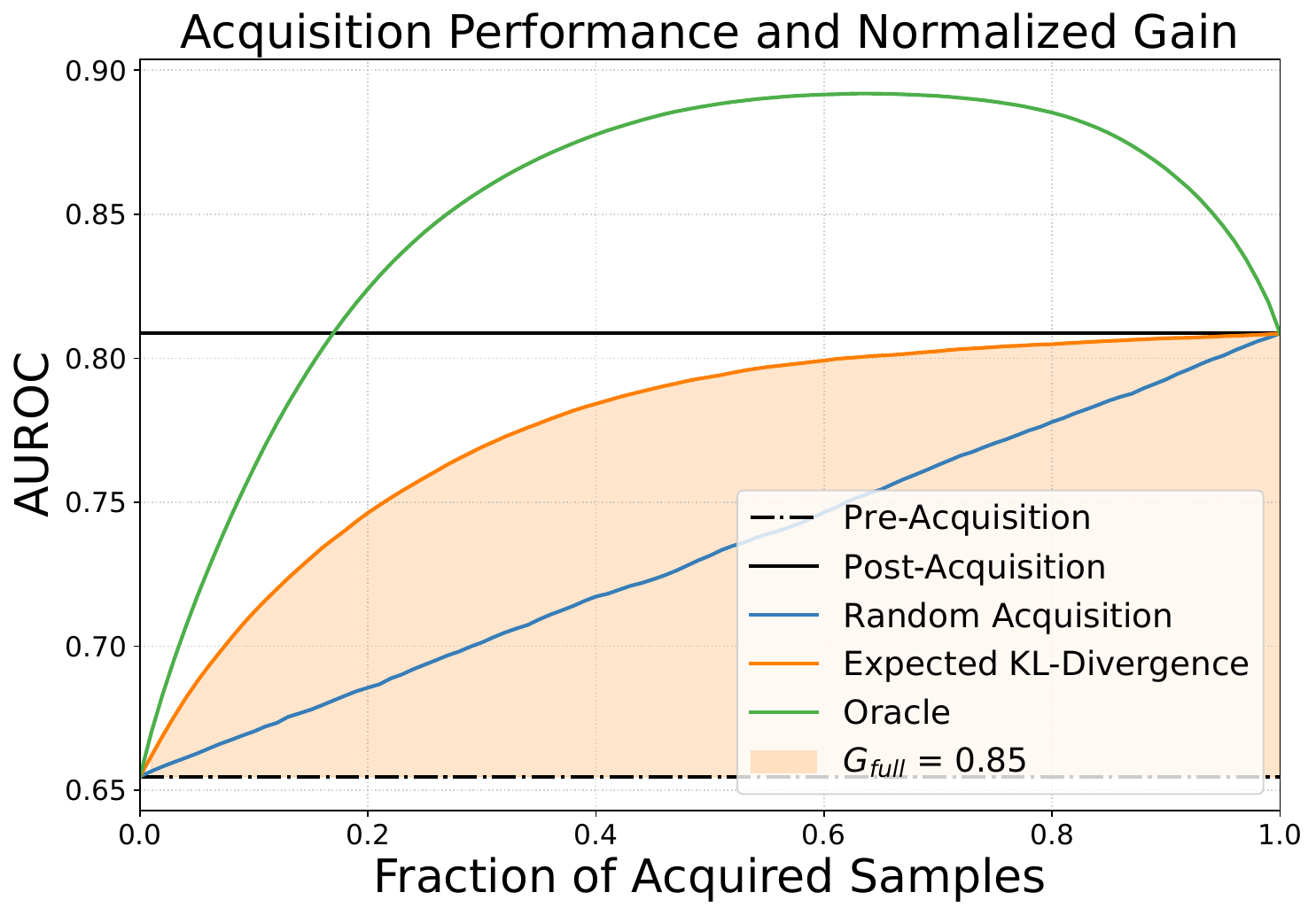}
        \caption{}
    \end{subfigure}

    \caption{
    \textbf{(a)} \ac{auroc} curves for \acp{af} on the MOSEI dataset at an acquisition budget of $25\%$ of the dataset size. \textbf{(b)} Gain achieved during the progressive acquisition of modalities. The oracle \ac{af} can exceed the post-acquisition \ac{auroc} before declining towards it again.
    }
    \label{fig:auroc_curves_side_by_side}
\end{figure}
\paragraph{Experimental Setup}
For datasets with at least three modalities, we apply five-fold cross-validation. Due to initially noisy results for MIMIC HAIM, we increase the number of folds to ten.
For each sample in the test set, the initial score $\rs_i^{\textnormal{avail}}$ is established by randomly assigning a subset of available modalities $\gP_i^{\textnormal{avail}}$. This procedure is repeated over several runs for robustness. In each run, every sample is stochastically assigned a new subset $\gP_i^{\textnormal{avail}}$. Acquisition is simulated by incrementally increasing the budget $\beta$. We focus on tasks where the post-acquisition model demonstrates a performance improvement over the pre-acquisition baseline since a simpler pre-acquisition model can outperform a more complex post-acquisition one, potentially due to the introduction of noisy or conflicting signals. In such cases, the final post-acquisition performance falls below the pre-acquisition baseline, resulting in a negative normalized area of gain. To ensure a meaningful evaluation, we exclude any tasks exhibiting this negative gain from the analysis at the split level.
For each budget, the top-ranked samples in $\gS$ are considered acquired, and their logits are updated from $\rs_i^{\textnormal{avail}}$ to $\rs_i^{\textnormal{acquired}}$. Reported results are aggregated across all cross-validation splits, combinations of missing and available modalities, and random runs to ensure robustness.
\paragraph{Results} 
\begin{table*}[t]
    \centering
    \scriptsize
    \caption{
        Acquisition performance on Symile, with $G_{\text{full}}$ shown for \ac{auroc}/\ac{auprc} as an example for the class with the best and worst performance and the mean value of all ten classes. Strategies are grouped by category. Best strategy among proposed ones and baselines in bold for each column. 
    }
    \label{tab:acquisition_strategy_performance_auroc_auprc_mimic_symile}
    \begin{tabular}{@{}l ccc ccc@{}}
        \toprule
        & \multicolumn{3}{c}{\textbf{Acquisitions by \ac{auroc}, $G_{\text{full}} \uparrow \pm \text{ \ac{sem}}$}} & \multicolumn{3}{c}{\textbf{Acquisitions by \ac{auprc}, $G_{\text{full}} \uparrow \pm \text{ \ac{sem}}$}} \\
        \cmidrule(lr){2-4} \cmidrule(lr){5-7}
        \textbf{Strategy} 
        & \textbf{Cardiomegaly} & \textbf{Pneumothorax} & \textbf{Mean}
        & \textbf{Lung Lesion} & \textbf{Pneumothorax} & \textbf{Mean} \\
        
        \midrule
        \multicolumn{7}{l}{\hspace{-6pt}\textit{Upper Bounds (for reference)}} \\
        Oracle & $2.787 \pm 0.139$ & $9.461 \pm 1.049$ & $4.580$ & $2.520 \pm 0.250$ & $10.623 \pm 0.708$ & $4.231$ \\
        True KL-Div. & $0.885 \pm 0.011$ & $0.910 \pm 0.054$ & $0.883$ & $0.828 \pm 0.073$ & $0.827 \pm 0.043$ & $0.871$ \\
        True Rank & $0.878 \pm 0.019$ & $0.605 \pm 0.053$ & $0.811$ & $0.676 \pm 0.088$ & $0.483 \pm 0.075$ & $0.776$ \\
        True Uncert. & $0.524 \pm 0.025$ & $-0.136 \pm 0.065$ & $0.481$ & $0.181 \pm 0.067$ & $0.293 \pm 0.052$ & $0.450$ \\

        \midrule 
        \multicolumn{7}{l}{\hspace{-6pt}\textit{Imputation-based (proposed)}} \\
        \ac{kldiv} & $\mathbf{0.747 \pm 0.039}$ & $0.773 \pm 0.134$ & $\mathbf{0.833}$ & $\mathbf{0.896 \pm 0.146}$ & $0.581 \pm 0.084$ & $\mathbf{0.777}$ \\
        Probability & $0.350 \pm 0.053$ & $\mathbf{0.898 \pm 0.061}$ & $0.426$ & $0.320 \pm 0.104$ & $\mathbf{0.965 \pm 0.027}$ & $0.449$ \\
        Rank & $0.378 \pm 0.016$ & $0.115 \pm 0.082$ & $0.378$ & $0.564 \pm 0.086$ & $0.396 \pm 0.054$ & $0.407$ \\
        Uncertainty & $0.450 \pm 0.041$ & $0.055 \pm 0.060$ & $0.440$ & $0.130 \pm 0.053$ & $0.513 \pm 0.066$ & $0.444$ \\

        \midrule 
        \multicolumn{7}{l}{\hspace{-6pt}\textit{Baselines (no imputation)}} \\
        Uncertainty & $0.480 \pm 0.013$ & $0.536 \pm 0.040$ & $0.480$ & $0.215 \pm 0.033$ & $0.811 \pm 0.041$ & $0.443$ \\
        Probability & $0.431 \pm 0.015$ & $0.536 \pm 0.040$ & $0.458$ & $0.756 \pm 0.136$ & $0.811 \pm 0.041$ & $0.550$ \\
        Random & $0.385 \pm 0.015$ & $0.327 \pm 0.061$ & $0.376$ & $0.503 \pm 0.103$ & $0.527 \pm 0.053$ & $0.388$ \\
        \bottomrule
    \end{tabular}
\end{table*}
\begin{table*}
    \begin{minipage}{.48\linewidth}
    \centering
    \scriptsize
    \caption{
        Acquisition performance on \ac{ukbb}. \acp{af} are grouped by category. Best strategy among proposed ones and baselines in bold.
    }
    \label{tab:results_ukbb_auroc_aucpr}
    \begin{tabular}{@{}l c c@{}}
        \toprule
        & \textbf{\ac{auroc}} & \textbf{\ac{auprc}} \\
        \cmidrule(lr){2-2} \cmidrule(lr){3-3}
        \textbf{Strategy} 
        & $G_{\text{full}} \pm \text{SEM} \uparrow$ & $G_{\text{full}} \pm \text{SEM} \uparrow$ \\
        \midrule
        \multicolumn{3}{l}{\hspace{-6pt}\textit{Upper Bounds (for reference)}} \\
        Oracle & $1.141 \pm 0.051$ & $1.721 \pm 0.315$ \\
        True KL-Div. & $0.978 \pm 0.007$ & $0.986 \pm 0.005$ \\
        True Rank & $0.887 \pm 0.022$ & $0.466 \pm 0.110$ \\
        True Uncert. & $0.436 \pm 0.088$ & $0.507 \pm 0.074$ \\
        \midrule
        \multicolumn{3}{l}{\hspace{-6pt}\textit{Imputation-based (proposed)}} \\
        \ac{kldiv} & $\mathbf{0.641 \pm 0.029}$ & $0.658 \pm 0.045$ \\
        Probability & $0.535 \pm 0.026$ & $\mathbf{0.713 \pm 0.029}$ \\
        Rank & $0.437 \pm 0.028$ & $0.340 \pm 0.114$ \\
        Uncertainty & $0.373 \pm 0.053$ & $0.332 \pm 0.058$ \\
        \midrule 
        \multicolumn{3}{l}{\hspace{-6pt}\textit{Baselines (no imputation)}} \\
        Uncertainty & $0.365 \pm 0.042$ & $0.556 \pm 0.073$ \\ 
        Probability & $0.365 \pm 0.042$ & $0.556 \pm 0.073$ \\ 
        Random & $0.528 \pm 0.018$ & $0.485 \pm 0.052$ \\
        \bottomrule
    \end{tabular}
    \end{minipage}
    \hfill 
    \begin{minipage}{.48\linewidth} 
    \centering
    \scriptsize
    \caption{
        Acquisition performance on MOSEI. \acp{af} are grouped by category. Best strategy among proposed ones and baselines in bold.
    }
    \label{tab:acquisition_strategy_performance_auroc_auprc_mosei}
    \begin{tabular}{@{}l c c@{}}
        \toprule
        & \textbf{\ac{auroc}} & \textbf{\ac{auprc}} \\
        \cmidrule(lr){2-2} \cmidrule(lr){3-3}
        \textbf{Strategy} 
        & $G_{\text{full}} \pm \text{SEM} \uparrow$ & $G_{\text{full}} \pm \text{SEM} \uparrow$ \\
        \midrule
        \multicolumn{3}{l}{\hspace{-6pt}\textit{Upper Bounds (for reference)}} \\
        Oracle & $1.478 \pm 0.091$ & $1.666 \pm 0.161$ \\
        True KL-Div. & $0.882 \pm 0.006$ & $0.838 \pm 0.006$ \\
        True Rank & $0.849 \pm 0.008$ & $0.806 \pm 0.010$ \\
        True Uncert. & $0.663 \pm 0.006$ & $0.708 \pm 0.005$ \\
        \midrule 
        \multicolumn{3}{l}{\hspace{-6pt}\textit{Imputation-based (proposed)}} \\
        \ac{kldiv} & $\mathbf{0.855 \pm 0.034}$ & $\mathbf{0.889 \pm 0.052}$ \\
        Probability & $0.707 \pm 0.037$ & $0.846 \pm 0.070$ \\
        Rank & $0.432 \pm 0.014$ & $0.457 \pm 0.019$ \\
        Uncertainty & $0.630 \pm 0.015$ & $0.706 \pm 0.037$ \\
        \midrule 
        \multicolumn{3}{l}{\hspace{-6pt}\textit{Baselines (no imputation)}} \\
        Uncertainty & $0.525 \pm 0.005$ & $0.540 \pm 0.006$ \\ 
        Probability & $0.433 \pm 0.007$ & $0.543 \pm 0.009$ \\ 
        Random & $0.490 \pm 0.004$ & $0.525 \pm 0.003$ \\
        \bottomrule
    \end{tabular}
    \end{minipage}
\end{table*}

In the following, we show the effectiveness of \ac{cama} and the proposed model architecture with downstream tasks (\Cref{fig:auroc_curves_side_by_side}, \Cref{tab:acquisition_strategy_performance_auroc_auprc_mimic_symile,tab:acquisition_strategy_performance_auroc_auprc_mosei,tab:results_ukbb_auroc_aucpr}), ablations (\Cref{tab:ablation}) and the general performance added due to modality acquisitions (\Cref{tab:auroc_binary_and_multiclass_mean}). 
First, as a sanity check and as expected, the \ac{auroc} on every dataset is increased after acquisition of the additional modality (\Cref{tab:auroc_binary_and_multiclass_mean}).
Second, oracle \acp{af} serve as an upper bound and achieve the highest performance (\Cref{tab:acquisition_strategy_performance_auroc_auprc_mimic_symile,tab:acquisition_strategy_performance_auroc_auprc_mosei,tab:results_ukbb_auroc_aucpr}). Surprisingly, oracle gains can exceed the value of one, as a strategic mix of pre-acquisition and post-acquisition samples can outperform a purely post-acquisition cohort. To benchmark the acquisition logic itself, we use label-agnostic upper-bound heuristics that access acquired scores $\rs_i^{\textnormal{acquired}}$.\begin{wraptable}[9]{r}{0.48\textwidth}
%\begin{table}[t]
    \centering
    \small
    \vspace{-0.42cm}
    \caption{Pre- and post-acquisition AUROC (Details in \Cref{app:model_performance_detailed}).}
    \label{tab:auroc_binary_and_multiclass_mean}
    \begin{tabular}{lcc}
        \toprule
        \textbf{Dataset} & \textbf{$M_\text{pre}$} & \textbf{$M_\text{post}$} \\
        \midrule
        MOSEI & $0.6500 \pm 0.0633$ & $0.8009 \pm 0.0060$ \\
        UKB & $0.5967 \pm 0.0745$ & $0.7214 \pm 0.0830$ \\
        Symile & $0.5924 \pm 0.0310$ & $0.6221 \pm 0.0265$ \\
        HAIM & $0.6783 \pm 0.0234$ & $0.6865 \pm 0.0196$ \\
        \bottomrule
    \end{tabular}
%\end{table}
\end{wraptable}\acp{af} based on \ac{kldiv} and rank change perform well, indicating that prioritizing large predictive shifts or cohort reordering is an effective heuristic in this setting.
The imputation-based \ac{kldiv} strategy consistently outperforms all other non-oracle methods. This \ac{af} effectively identifies samples predicted to have the largest shift in their class probability distribution (\Cref{fig:auroc_curves_side_by_side}). In contrast, imputation-based \acp{af} relying on rank change, final uncertainty, or final probability are considerably weaker, suggesting that the change in prediction is more effective than estimating the final state. While our primary results for imputation-based \acp{af} use \acp{ddpm}, a \ac{bcvae} variant offers faster inference for a trade-off in performance (\Cref{tab:runtime,app:rbt_bcvae_symile}).
Overall, this confirms the robustness and scalability of our framework in a challenging setting. Our results affirm the superiority of the imputation-based \ac{kldiv} strategy, which achieved substantial and reliable gains over all baselines and heuristics. Details in \Cref{app:detailed_results_mosei,app:detailed_results_mimic_symile,app:results_mimic_haim}.

\section{Discussion}
\label{sec:discussion}
\begin{wraptable}[9]{r}{0.43\textwidth}
    \centering
    \small
    \vspace{-0.45cm}
    \caption{Cross-validated ablation on the Symile dataset.}
    \label{tab:ablation}
    \begin{tabular}{@{}p{4.6cm}c@{}}
        \toprule
        \textbf{Ablation (acquisitions by AUROC)} & \textbf{$\text{G}_{\text{full}}$ $\uparrow$} \\
        \midrule
        \ac{kldiv} (w.r.t.\ \Cref{tab:acquisition_strategy_performance_auroc_auprc_mimic_symile}) & 0.833 \\
        \textit{w/o} Layer Norm & 0.772 \\
        \textit{w/o} label smoothing & 0.746 \\
        \textit{w/o} decoupled data flow & 0.599 \\
        \textit{w/o} balanced train set & 0.568 \\
        \bottomrule
    \end{tabular}
\end{wraptable}

% Introduction
We introduce \ac{cama} to address modality acquisition under budget constraints. While imputation-based \acp{af} provide an effective solution, we discuss the key implications in the following.
% Oracle Behavior 
\paragraph{Oracle Behavior}
Oracles can exceed a model using post-acquisition data for all samples (\Cref{fig:auroc_curves_side_by_side} (b))\ie an underlying predictive model can achieve better global performance with a strategic curation of samples. One hypothesis is that additional modalities may introduce variance, redundancy, or conflicting information that imperfect models cannot reconcile but the oracles can circumvent.
% KL > upper bound, gap between rank-heuristic and imputation-heuristic
Surprisingly, the imputation-based \ac{kldiv} \ac{af} can outperform the corresponding upper-bound heuristic (\Cref{tab:acquisition_strategy_performance_auroc_auprc_mosei}). Conversely, the performance gap between the rank-change heuristic and its imputation-based counterpart suggests that global, rank-based metrics may be vulnerable to imputation noise. While the \ac{kldiv} \ac{af} demonstrated strong performance, not all imputation-based \acp{af} outperformed simpler \acp{af} (\Cref{app:detailed_results_mimic_symile,app:results_mimic_haim}). This indicates that optimal \acp{af} can be context-dependent and that imputation quality is not the only factor.
\paragraph{Imputation Quality and Modality Relationships}
We impute latent embeddings optimized for the discriminative task rather than raw data (\Cref{fig:model_architecture}). Consequently, standard imputation metrics\eg \ac{fid}, are not applicable across different generative models. Every generative model influences the encoders latent spaces indirectly. While we observe that utilizing stronger generative models results in higher acquisition performance, generative imputation quality is not the only factor (\Cref{tab:ablation}). 
Further, the relationship between imputation quality and the degree of shared information between modalities is important. Intuitively, imputation is expected to be easier when the target modality is more strongly predictable from the observed modalities. 
We show \ac{cama}s robustness to imputation errors since $f_{\text{imp}}$ models a distribution of plausible outcomes rather than aiming for a single reconstruction. By averaging the expected impact across this distribution, the acquisition decision becomes less sensitive to uncertainty. 
%Additionally, the primary \ac{kldiv} \ac{af} is resilient to noise, as it prioritizes samples expected to cause a large predictive shift, effectively ignoring minor imputation errors. 
% Big picture
%\ac{cama} is not only practical for constrained settings, but also reveals insights into post-acquisition behavior. The successful \ac{kldiv} strategy and the surprising oracle performance underscore that the value of an additional modality is not absolute but highly contextual. The most effective \acp{af} are not those that simply predict an outcome, but estimate the magnitude of the predictive shift.
% Limitations
\paragraph{Limitations}
We study acquisition of a single specified modality under a fixed cohort-level budget, which is meaningful for resource-specific settings (\Cref{sec:introduction}) but does not address the more general problem of jointly selecting both the modality and the sample (\Cref{app:multimodal_acquisition}). We use randomized missingness patterns as a controlled setup, which facilitates comparison across datasets but does not capture the missingness found in real-world applications. We also do not model heterogeneous acquisition costs across modalities or individuals (\Cref{app:multimodal_acquisition}). Finally, we do not include a dedicated fairness analysis which may lead to further interesting insights.
\section{Conclusion and Future Work}
\label{sec:conclusion}
We introduce \ac{cama}, a novel setting addressing the real-world challenge of optimizing global discriminative performance through strategic test-time acquisition of an additional modality under resource constraints. Our evaluation across multiple multimodal datasets shows that imputation-based \acp{af} can effectively guide resource allocation under cohort-level constraints. In settings such as healthcare, strategic allocation of costly or invasive diagnostic procedures is essential, and our approach offers a promising direction for these applications.
Future work includes extending \ac{cama} to multi-class problems or regression tasks, exploring additional imputation techniques, and dynamically selecting which modality to acquire instead of pre-selecting one.
\clearpage

\paragraph{Acknowledgement}
%\label{app:acknowledgement}
We would like to thank Stefan Hegselmann and Lucas Arnoldt for the supportive and insightful discussions throughout the research and development process. The authors acknowledge the Scientific Computing of the IT Division at the Charité - Universitätsmedizin Berlin for providing computational resources that have contributed to the research results reported in this paper. This research has been conducted using the UK Biobank Resource under application number 49966. B.W. and R.E. acknowledge support by the Collaborative Research Center (SFB 1470) funded by the German Research Council (DFG).

\bibliography{references}
\bibliographystyle{plainnat}

\newpage 
\appendix
\section{Broader Impact and Ethics}
\label{app:broader_impact}
The \ac{cama} setting introduced in this paper offers potential for positive broader impacts, primarily by enabling more efficient use of resources in multimodal machine learning. In resource-constrained fields like healthcare, this could facilitate access to more robust and comprehensive model performance by strategically guiding the acquisition of costly or limited additional data modalities. This could translate to improved diagnostic accuracy where such data is critical but not uniformly available for all samples in a cohort.
However, the deployment of \ac{cama}, particularly its core function of ranking and prioritizing samples for modality acquisition, necessitates careful ethical consideration. This raises concerns about equity and fairness, especially if the downstream application impacts critical decisions. A significant risk is the potential to introduce biases, including racial, socioeconomic, or other demographic biases. Therefore, the development and application of \ac{cama} must be approached with a strong commitment to ethical principles.

\section{Reproducibility}
\label{app:reproducibility_statement}
To ensure the reproducibility of our results, we provide the following details:

\paragraph{Code} The complete source code used for all experiments will be made publicly available on GitHub upon publication. The repository will include scripts for model training and evaluation.
    
\paragraph{Hyperparameters} All hyperparameters, including learning rates, batch sizes, and model-specific parameters, are explicitly listed in \Cref{app:hyperparameter_sweeps}. Additionally, we provide the complete sweep configurations used for hyperparameter tuning to allow for full replication of our optimization process.
    
\paragraph{Datasets} Three of the four datasets used in our evaluation are publicly available. For more details see \Cref{sec:evaluation} and \Cref{app:dataset_details}.

\paragraph{Implementation Details} We provide a full section in \Cref{app:hyperparameter_sweeps} and a dedicated paragraph in \Cref{sec:evaluation} describing implementation details that we found to be crucial.

\section{Details about Acquisition Function Strategies}
\label{app:details_about_acquisition_function_strategies}

\subsection{AUROC and AUPRC}
\label{subsec:auroc_and_auprc}
To derive the proposed acquisition strategies, we briefly explain the metrics used in the following paragraphs.

\paragraph{\ac{auroc}} The \acf{auroc} measures the model's ability to discriminate between positive and negative classes and is defined as

\begin{equation}
    \text{AUROC}(\rvy, \rvs) = \frac{1}{N_+ N_-} \sum_{i : \ry_i=1} \sum_{j : \ry_j=0} \left( \indicator(\rs_i > \rs_j) + \frac{1}{2}\indicator(\rs_i = \rs_j) \right)
    \label{eq:auroc}
\end{equation}

where $N_+ = |\{i \mid \ry_i=1\}|$ and $N_- = |\{j \mid \ry_j=0\}|$.

\paragraph{\ac{auprc}} The \acf{auprc} summarizes the trade-off between precision ($P_t$) and recall ($R_t$) across different decision thresholds $t$ and is defined as

\begin{align}
    \text{AUPRC}(\rvy, \rvp) = \sum_{k=1}^{N'} (R_k - R_{k-1}) P_k
    \label{eq:auprc}
\end{align}

where points $(R_k, P_k)$ are ordered by threshold from the PR curve, $N'$ is the number of unique thresholds, and $\rvp = \sigmoid(\rvs)$.

\subsection{Oracle Acquisition Strategies: Exact Gain Calculation}
\label{subsec:oracle_strategies}

Oracle acquisition strategies serve as theoretical upper limits for the performance of greedy acquisition approaches. They operate under the ideal assumption that the true labels $\ry_i$ and the outcome scores $\rs_i^{\textnormal{acquired}}$ are known for all samples $i \in \{1, \dots, N\}$. While not implementable in practice, these oracle strategies provide benchmarks by selecting samples based on their exact marginal contribution to the global evaluation metric. The general principle is to iteratively select $\beta$ samples. At each step, among the samples for which the additional modality has not yet been acquired, the oracle picks the one that provides the largest true immediate gain to the chosen global metric.

\paragraph{\ac{auroc} Oracle}
The \ac{auroc} oracle strategy aims to maximize the cohort's \ac{auroc} by identifying, at each step, the sample $i$ that yields the largest immediate increase in this metric if its additional modality were acquired (changing its score from $\rs_i^{\textnormal{avail}}$ to $\rs_i^{\textnormal{acquired}}$)\ie a greedy selection. This prospective increase is quantified by the marginal gain $g_i^{\text{\ac{auroc}}}$. The components of this gain, $g_i^{\text{\ac{auroc}}}(\ry_i=1)$ (for positive samples) and $g_i^{\text{\ac{auroc}}}(\ry_i=0)$ (for negative samples), reflect the net change in favorable pairwise score comparisons relative to samples of the other class. Recall the definition of \ac{auroc} from \Cref{eq:auroc}:

\begin{equation*}
    \text{AUROC}(\rvy, \rvs) = \frac{1}{N_+ N_-} \sum_{i : \ry_i=1} \sum_{j : \ry_j=0} \left( \indicator(\rs_i > \rs_j) + \frac{1}{2}\indicator(\rs_i = \rs_j) \right).
\end{equation*}

The total marginal gain for sample $i$, representing the exact change in the cohort's \ac{auroc} value, is then, by considering positive and negative samples and neglecting the normalization factor:

%\begin{equation}
%   g_i^{\text{\ac{auroc}}}(\ry_i=1) = \sum_{j:\ry_j=0} \left( \indicator(\rs_i^{\textnormal{full}} > \rs_j^{\textnormal{avail}}) - \indicator(\rs_i^{\textnormal{avail}} > \rs_j^{\textnormal{avail}}) + \frac{1}{2}\left[\indicator(\rs_i^{\textnormal{full}} = \rs_j^{\textnormal{avail}}) - \indicator(\rs_i^{\textnormal{avail}} = \rs_j^{\textnormal{avail}})\right] \right)
%    \label{eq:auroc_gain_positive}
%\end{equation}

%\begin{equation}
%    g_i^{\text{\ac{auroc}}}(\ry_i=0) = \sum_{j:\ry_j=1} \left( \indicator(\rs_j^{\textnormal{avail}} > \rs_i^{\textnormal{full}}) - \indicator(\rs_j^{\textnormal{avail}} > \rs_i^{\textnormal{avail}}) + \frac{1}{2}\left[\indicator(\rs_j^{\textnormal{avail}} = \rs_i^{\textnormal{full}}) - \indicator(\rs_j^{\textnormal{avail}} = \rs_i^{\textnormal{avail}})\right] \right)
%    \label{eq:auroc_gain_negative}
%\end{equation}

\begin{equation}
\label{eq:auroc_gain_positive}
\begin{split}
    g_i^{\text{\ac{auroc}}}(\ry_i=1) &= \sum_{j:\ry_j=0} \bigg( \indicator(\rs_i^{\textnormal{acquired}} > \rs_j^{\textnormal{avail}}) - \indicator(\rs_i^{\textnormal{avail}} > \rs_j^{\textnormal{avail}}) \\
    &\quad + \frac{1}{2}\Big[\indicator(\rs_i^{\textnormal{acquired}} = \rs_j^{\textnormal{avail}}) - \indicator(\rs_i^{\textnormal{avail}} = \rs_j^{\textnormal{avail}})\Big] \bigg)
\end{split}
\end{equation}

\begin{equation}
\label{eq:auroc_gain_negative}
\begin{split}
    g_i^{\text{\ac{auroc}}}(\ry_i=0) &= \sum_{j:\ry_j=1} \bigg( \indicator(\rs_j^{\textnormal{avail}} > \rs_i^{\textnormal{acquired}}) - \indicator(\rs_j^{\textnormal{avail}} > \rs_i^{\textnormal{avail}}) \\
    &\quad + \frac{1}{2}\Big[\indicator(\rs_j^{\textnormal{avail}} = \rs_i^{\textnormal{acquired}}) - \indicator(\rs_j^{\textnormal{avail}} = \rs_i^{\textnormal{avail}})\Big] \bigg)
\end{split}
\end{equation}

\begin{equation}
    g_i^{\text{\ac{auroc}}} = \frac{1}{N_+ N_-} \left( g_i^{\text{\ac{auroc}}}(\ry_i=1) \cdot \indicator(\ry_i=1) + g_i^{\text{\ac{auroc}}}(\ry_i=0) \cdot \indicator(\ry_i=0) \right)
    \label{eq:auroc_gain_total}
\end{equation}

\paragraph{\ac{auprc} Oracle}
The \ac{auprc} oracle strategy seeks to maximize the cohort's \ac{auprc}. It operates by identifying, at each step, the sample $i$ which, if its additional modality were acquired (changing its score from $\rs_i^{\textnormal{avail}}$ to $\rs_i^{\textnormal{acquired}}$), would yield the largest immediate increase in the global \ac{auprc} value\ie a greedy selection. This marginal gain, $g_i^{\text{\ac{auprc}}}$, represents the exact change in the cohort's \ac{auprc}. 
To calculate the marginal gain for a sample $i$, we compute the change in the cohort's AUPRC. Let $\rvs^{\textnormal{current}}$ be the vector of scores for the whole cohort. We define a new vector, $\rvs^{\textnormal{updated}}$, which is identical to $\rvs^{\textnormal{current}}$ except that for sample i, the score is changed from $\rs_i^{\textnormal{avail}}$ to $\rs_i^{\textnormal{acquired}}$. The marginal gain is then:

\begin{equation}
g_i^{\text{\ac{auprc}}} = \text{\ac{auprc}} (\rvy, \rvp^{\textnormal{updated}}) - \text{\ac{auprc}} (\rvy, \rvp^{\textnormal{current}})
\end{equation}

where $\rvp^{\textnormal{current}} = \sigmoid(\rvs^{\textnormal{current}})$ and $\rvp^{\textnormal{updated}} = \sigmoid(\rvs^{\textnormal{updated}})$.

\subsection{Upper-Bound Heuristic Strategies}
The preceding oracle strategies make the assumption of perfect foresight into both the true labels $\ry_i$ and the exact outcome scores $\rs_i^{\textnormal{acquired}}$. We now introduce a distinct class of upper-bound heuristic strategies. These strategies still presume access to the true future scores $\rs_i^{\textnormal{acquired}}$ for any sample $i$ if its additional modality were acquired. However, the following upper-bound heuristics are label-agnostic\ie the true label $\ry_i$ of a candidate sample is not used when determining its priority for acquisition. Consequently, the selection principle for these strategies must rely on how the known change from an initial score $\rs_i^{\textnormal{avail}}$ to the future score $\rs_i^{\textnormal{acquired}}$ is expected to influence the global evaluation metric, without direct reference to the sample's ground-truth label.

\paragraph{Maximum True Uncertainty Reduction}
The uncertainty reduction strategy prioritizes acquiring the additional modality for samples where doing so is expected to yield the largest decrease in predictive uncertainty. For each sample $i$, uncertainty is quantified using the binary entropy $\entropy(p_i)$ of its predicted probability $p_i$ for the positive class, defined as:

\begin{equation}
    \entropy(p_i) = -p_i \log_2 p_i - (1-p_i) \log_2 (1-p_i),
    \label{eq:binary_entropy}
\end{equation}

The acquisition strategy operates with knowledge of the initial probability $p_i^{\textnormal{avail}} = \sigmoid(\rs_i^{\textnormal{avail}})$ derived from the available modalities, and crucially, the true future probability $p_i^{\textnormal{acquired}} = \sigmoid(\rs_i^{\textnormal{acquired}})$ that would be obtained if the additional modality were acquired (where $\rs_i^{\textnormal{acquired}}$ is the oracle score). The acquisition score $g_i^{\text{UR}}$ for sample $i$ is then the exact reduction in entropy:

\begin{equation}
    g_i^{\text{UR}} = \entropy(p_i^{\textnormal{avail}}) - \entropy(p_i^{\textnormal{acquired}}).
    \label{eq:UR_gain}
\end{equation}

Samples with higher $g_i^{\text{UR}}$ values, indicating a greater expected reduction in uncertainty, are prioritized for modality acquisition.

\paragraph{Maximum True Rank Change}
This rank change strategy prioritizes samples whose relative standing within the cohort, based on predicted probability of belonging to the positive class, would change most significantly if the additional modality were acquired. For each sample $i$, we consider its rank $R(p_i)$ when all $N$ samples in the cohort are ordered by their respective probabilities $p_i$. The acquisition score $g_i^{\text{RC}}$ for sample $i$ is defined as the absolute magnitude of this change in rank:

\begin{equation}
    g_i^{\text{RC}} = |R(p_i^{\textnormal{acquired}}) - R(p_i^{\textnormal{avail}})|.
\end{equation}

Samples exhibiting a higher $g_i^{\text{RC}}$ are prioritized for modality acquisition, since they are expected to cause the largest shift in the sample's rank-ordered position relative to its peers.

\paragraph{\ac{kldiv}}
The \ac{kldiv} acquisition strategy aims to identify samples for which acquiring the additional modality would lead to the largest change in the predicted probability distribution. Specifically, it quantifies the divergence from the predicted probability distribution based on the true future score, $P_i^{\textnormal{acquired}} \sim \text{Bernoulli}(p_i^{\textnormal{acquired}})$, back to the initial distribution based on baseline data, $P_i^{\textnormal{avail}} \sim \text{Bernoulli}(p_i^{\textnormal{avail}})$. This is measured by the \ac{kldiv} $\KLdiv(P_i^{\textnormal{avail}} \Vert P_i^{\textnormal{acquired}})$ and can be defined as follows for an acquisition function:

\begin{align}
    g_i^{\text{KLD}} 
    &= \KLdiv\left(P_i^{\textnormal{avail}} \middle\Vert P_i^{\textnormal{acquired}}\right) \\
    &= p_i^{\textnormal{avail}} \log_2 \frac{p_i^{\textnormal{avail}}}{p_i^{\textnormal{acquired}}} + (1-p_i^{\textnormal{avail}}) \log_2 \frac{1-p_i^{\textnormal{avail}}}{1-p_i^{\textnormal{acquired}}}
    \label{eq:kldiv_gain}
\end{align}

Samples with a higher $g_i^{\text{KLD}}$ are prioritized, as this indicates a greater discrepancy between the prediction based on available data and the prediction that would be made with the additional modality.

\subsection{Baseline Information Strategies}
Shifting from approaches that leverage oracle knowledge of future scores ($\rs_i^{\textnormal{acquired}}$), the present section details methods serving as practical, label-agnostic baselines. They make acquisition decisions based exclusively on information derived from the initially available modality ($\rs_i^{\textnormal{avail}}$). A random acquisition strategy serves as a fundamental baseline.

\paragraph{Maximum Baseline Uncertainty}
The Maximum Baseline Uncertainty strategy is a baseline that prioritizes samples for which the prediction based on the initially available modality is most uncertain. The acquisition score for sample $i$ is directly the binary entropy $\entropy(p_i^{\textnormal{avail}})$, as defined in \Cref{eq:binary_entropy}:

\begin{equation}
    g_i^{\text{UU}} = \entropy(p_i^{\textnormal{avail}}).
    \label{eq:uu_gain}
\end{equation}

Samples with a higher $g_i^{\text{UU}}$\ie $p_i^{\textnormal{avail}}$ closer to $0.5$, since the entropy $H(p_i^{\textnormal{avail}})$ is symmetric around $p_i^{\textnormal{avail}}=0.5$, are selected first.

\paragraph{Maximum Baseline Probability}
This approach prioritizes acquiring the additional modality for samples that the baseline model already predicts as belonging to the positive class with high confidence. The acquisition score $g_i^{\text{UP}}$ for sample $i$ is simply its initial probability $p_i^{\textnormal{avail}}$ based on the available modality:

\begin{equation}
    g_i^{\text{UP}} = p_i^{\textnormal{avail}},
    \label{eq:up_gain}
\end{equation}

Samples with a higher $g_i^{\text{UP}}$ are prioritized for acquisition.

\subsection{Imputation-Based Strategies}
Having explored strategies that assume perfect knowledge of the true labels $\ry_i$ and/or future scores $\rs_i^{\textnormal{acquired}}$, and simpler baselines relying only on current information $\rs_i^{\textnormal{avail}}$, we now introduce methods aiming to bridge the gap by offering a practical and label-agnostic pathway to modality acquisition. They operate by utilizing an imputation model, $f_{\text{imp}}$, to generate a set of $K$ plausible future scores, denoted $\{\rs_{i,k}^{\textnormal{imp}}\}_{k=1}^K$, conditioned on the initially available data $\rs_i^{\textnormal{avail}}$. The core principle of these strategies is to then derive acquisition scores from statistics of this imputed score distribution, with the goal of emulating the decision-making process, but without requiring true future knowledge at test time.

\paragraph{Maximum Expected Probability}
The Maximum Expected Probability strategy prioritizes samples which have the highest average probability of belonging to the positive class after modality acquisition. It relies on the set of $K$ imputed future probabilities $\{\rp_{i,k}^{\textnormal{imp}}\}_{k=1}^K$, where each $\rp_{i,k}^{\textnormal{imp}} = \sigmoid(\rs_{i,k}^{\textnormal{imp}})$ is derived from an imputed future score $\rs_{i,k}^{\textnormal{imp}}$. The acquisition score $g_i^{\text{eP}}$ for sample $i$ is the mean of these imputed probabilities:

\begin{equation}
    g_i^{\text{eP}} = \frac{1}{K} \sum_{k=1}^{K} \rp_{i,k}^{\textnormal{imp}}.
    \label{eq:ep_gain}
\end{equation}

Samples with a higher $g_i^{\text{eP}}$ are selected, representing instances where the imputation model, on average, predicts a high likelihood of being positive if the additional modality were acquired.

\paragraph{Maximum Expected Uncertainty Reduction}
The Maximum Expected Uncertainty Reduction strategy aims to select samples for which the acquisition of the additional modality is anticipated to yield the largest average decrease in predictive uncertainty (\Cref{eq:binary_entropy}). This strategy considers the initial entropy $\entropy(p_i^{\textnormal{avail}})$, and the distribution of entropies $\{\entropy(\rp_{i,k}^{\textnormal{imp}})\}_{k=1}^K$. The acquisition score $g_i^{\text{eUR}}$ is the difference between the initial entropy and the mean of the imputed future entropies:

\begin{equation}
    g_i^{\text{eUR}} = \entropy(p_i^{\textnormal{avail}}) - \frac{1}{K} \sum_{k=1}^{K} \entropy(\rp_{i,k}^{\textnormal{imp}}).
    \label{eq:eur_gain}
\end{equation}

Samples with higher $g_i^{\text{eUR}}$ are prioritized, indicating a greater expected clarification of the prediction upon acquiring the new modality.

\paragraph{Expected Rank Change}
The Maximum Expected Rank Change strategy prioritizes samples for which the acquisition of the additional modality is anticipated to cause the largest change in their rank, relative to the initial ranking based on $p_i^{\textnormal{avail}}$. It aims to mirror the "Maximum True Rank Change" strategy by using imputed future probabilities.
Let $R(p_i^{\textnormal{avail}})$ denote the rank of sample $i$ when all $N$ samples in the cohort are ordered by their initial probabilities $p_j^{\textnormal{avail}}$ (for $j=1, \dots, N$). For each of the $K$ imputed future probabilities $\rp_{i,k}^{\textnormal{imp}}$ for sample $i$, let $R(\rp_{i,k}^{\textnormal{imp}})$ denote the rank of sample $i$ if its probability were $\rp_{i,k}^{\textnormal{imp}}$ while all other samples $j \neq i$ retain their initial probabilities $p_j^{\textnormal{avail}}$. The acquisition score $g_i^{\text{eRC}}$ is then the mean of the absolute differences between these imputed future ranks and the initial rank:

\begin{equation}
    g_i^{\text{eRC}} = \frac{1}{K} \sum_{k=1}^{K} |R(\rp_{i,k}^{\textnormal{imp}}) - R(p_i^{\textnormal{avail}})|.
    \label{eq:merc_gain}
\end{equation}

Samples with a higher $g_i^{\text{eRC}}$ are selected, as they are expected to experience the largest shift in their rank-ordered position relative to other samples in the cohort upon modality acquisition.

\paragraph{Expected \ac{kldiv}}
The Expected \ac{kldiv} strategy selects samples where the initial probability distribution is expected to diverge most significantly from the future probability distributions derived from the $K$ imputed scores. The acquisition score $g_i^{\text{eKLD}}$ is the average \ac{kldiv} $\KLdiv(P_i^{\textnormal{avail}} \Vert P_i^{(\textnormal{imp},k)})$ over the $K$ imputations:

\begin{equation}
    g_i^{\text{eKLD}} = \frac{1}{K} \sum_{k=1}^{K} \KLdiv\left(P_i^{\textnormal{avail}} \middle\Vert P_i^{(\textnormal{imp},k)}\right).
    \label{eq:ekld_gain}
\end{equation}

A higher $g_i^{\text{eKLD}}$ indicates that, on average, the imputed future predictions substantially differ from the initial baseline prediction, suggesting a significant informational update from acquiring the additional modality.

\section{Hyperparameters, Model Details and Compute Environment}
\label{app:hyperparameter_sweeps}
We employ domain-specific encoders to process the respective modalities: for language inputs, we use a pre-trained BERT model \citep{devlin_bert_2019}, for vision, a Vision Transformer (ViT) \citep{dosovitskiy_image_2021}. Other data types\eg temporal sequences, tabular data, or pre-extracted embeddings, are handled by Transformer encoders \citep{vaswani_attention_2017}.
We use well-established hyperparameters from the literature for the modality-specific encoders and only optimize the remaining parameters. Notably, our experiments compared three approaches for normalizing the encoder output: No Normalization, Batch Normalization, and Layer Normalization. We found Layer Normalization to be particularly advantageous, as it both stabilized training convergence and significantly enhanced the performance of the \acp{ddpm}. We also evaluated the impact of using only the \texttt{CLS} token representation from the encoder versus leveraging the full output sequence. This comparison revealed no substantial effect on performance, suggesting the sufficiency of the \texttt{CLS} token representation for our task. Layers in the network are initialized using He initialization \citep{he_delving_2015} if they were not pre-initialized by the specific encoder architecture. We find this particularly important for stabilizing the \acp{ddpm} during the early epochs of end-to-end model training.

We perform hyperparameter sweeps for the remaining parts of the designed model in the following ranges:

\begin{itemize}
    \item Transformer Head
    \SubItem {Embedding dimension: \texttt{[32, 64, 128, 256, 512, 1024]}}
    \SubItem {Feed-Forward network: \texttt{[128, 256, 512, 1024, 2048]}}
    \SubItem {Dropout: \texttt{[0, 0.1, 0.2]}}
    \SubItem {Number of heads: \texttt{[4, 8, 16]}}
    \SubItem {Number of layers: \texttt{[2, 4, 6, 8]}}

    \item \acp{ddpm}
    \SubItem {Embedding dimension: analogous to Transformer head}
    \SubItem {Hidden dimension: \texttt{[32, 64, 128, 256, 512, 1024]}}
    \SubItem {Dropout: \texttt{[0, 0.1, 0.2]}}
    \SubItem {Number of heads: \texttt{[4, 8, 16]}}
    \SubItem {Number of layers: \texttt{[2, 4, 6, 8]}}
    \SubItem {Number of steps: \texttt{[10, 25, 50, 100, 250, 500]}}

    \item ScheduleFree Optimizer
    \SubItem {Learning rate: \texttt{[1e-1, 1e-2, 1e-3, 3e-4, 1e-4, 1e-5]}}
    \SubItem {Warmup steps: \texttt{[0, 100, 200]}}
    \SubItem {Weight decay: \texttt{[0, 0.01, 0.001]}}
    
\end{itemize}

The models are trained with early stopping but without any maximum number of epochs. For the imputation-based acquisition functions, $100$ \ac{ddpm} samples are used during inference of the model.

\begin{figure}  %[16]{R}{0.4\textwidth}  %
    \centering
    \includegraphics[width=0.4\textwidth]{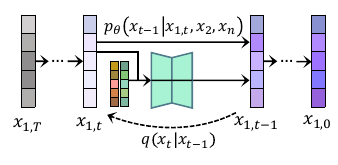}
    \caption{The latent \ac{ddpm} with its (de)noising functions. Coloring represents less noise in the latent space, starting with pure noise in $X_{i,T}=X_{1,T}$ with $T$ steps. The \ac{ddpm} is conditioned with two non-missing latent spaces, each from one remaining modality respectively.}
    \label{fig:latent_ddpm}
\end{figure}

Our experiments are conducted on a High-Performance Cluster (HPC) with the following environment:

\begin{enumerate}
    \item 21 Dell PowerEdge R7525 compute nodes, each with:
        \begin{itemize}
            \item 64 AMD Epyc cores (Rome)
            \item 512GB RAM
            \item 1 NVIDIA A100 40G GPU
        \end{itemize}
    \item 2 Dell PowerEdge XE8545 compute nodes, each with:
        \begin{itemize}
            \item 128 AMD Epyc cores (Milan)
            \item 512GB RAM
            \item 4 NVIDIA A100 40G GPUs (NVLink-connected)
        \end{itemize}
\end{enumerate}

\section{Dataset Details}
\label{app:dataset_details}
We evaluate \ac{cama} on four diverse, real-world multimodal datasets, spanning domains from healthcare to emotion recognition. 

\paragraph{MIMIC Symile}
This clinical dataset is derived from the MIMIC database and is designed for predicting the diagnosis of ten classes (Fracture, Enlarged Cardiomediastinum, Consolidation, Atelectasis, Edema, Cardiomegaly, Lung Lesion, Lung Opacity, Pneumonia, Pneumothorax). It contains 10,345 samples from patients in intensive care units. For our experiments, we utilize three distinct modalities: laboratory values, chest X-ray images, and \acp{ecg}.

\paragraph{MIMIC HAIM}
This healthcare benchmark also focuses on the diagnostic prediction of ten classes (Fracture, Enlarged Cardiomediastinum, Consolidation, Atelectasis, Edema, Cardiomegaly, Lung Lesion, Lung Opacity, Pneumonia, Pneumothorax). The bimodal dataset consists of 45,050 samples. The two modalities used in our study are laboratory values and chest X-ray images.

\paragraph{CMU-MOSEI}
This large-scale benchmark targets multimodal sentiment analysis and emotion recognition with seven classes covering different emotions. It contains 22,856 video samples of speakers expressing opinions. The dataset comprises three modalities: vision, acoustics, and language. Notably, unlike the other datasets, we utilize the pre-computed embeddings provided by the authors rather than the raw data.

\paragraph{UK Biobank (UKBB)}
The UK Biobank is a large-scale, prospective biomedical database from half a million UK participants. In our experiments, the costly modality targeted for acquisition is proteomics, which is available for only a fraction of the full cohort. We constructed a subset of 100,000 samples in which approximately half include proteomics data, accurately simulating a resource-constrained acquisition scenario. The 15 modalities utilized include \acp{ehr}, NMR metabolomics, proteomics, physical activity measurements, diet and alcohol consumption questionnaires, baseline characteristics, smoking status, physiological measurements, anthropometry, hand grip strength, cognitive function tests, \acp{ecg}, \ac{prs}, and arterial stiffness measurements.

\section{Extension to Flexible Modality Acquisition}
\label{app:multimodal_acquisition}
In the main paper, we study \ac{cama} in the setting where one specified missing modality is considered for acquisition under a fixed budget \(\beta\). A natural extension is to allow the acquisition policy to decide \emph{which} missing modality to acquire for each sample, or even to acquire multiple missing modalities per sample. We briefly outline this extension here to clarify that it can be built on top of the present framework, but introduces a substantially more complex optimization problem.

\paragraph{From Fixed-Modality Acquisition to Flexible Modality Selection}
In the setting considered in the main paper, the optimization problem is to select a subset of samples \(\gS \subseteq \{1,\dots,N\}\) with \(|\gS|=\beta\), where the same target modality is acquired for every selected sample. The resulting score for sample \(i\) is then
\[
\rs_i(\gS) =
\begin{cases}
\rs_i^{\textnormal{acquired}} & \text{if } i \in \gS \\
\rs_i^{\textnormal{avail}} & \text{if } i \notin \gS.
\end{cases}
\]
Thus, the acquisition functions only need to rank samples according to the expected utility of acquiring that one target modality.

In a flexible multi-modality setting, however, each sample \(i\) may have several missing modalities available for acquisition. If \(\gP_i^{\textnormal{avail}}\) denotes the set of currently available modalities, then the set of candidate missing modalities is
\[
\gP_i^{\textnormal{miss}} = \{1,\dots,P\} \setminus \gP_i^{\textnormal{avail}}.
\]
The acquisition decision is no longer only whether sample \(i\) should be selected, but also which modality \(p \in \gP_i^{\textnormal{miss}}\) should be acquired. The decision space therefore changes from ranking samples \(i\) to ranking sample-modality pairs \((i,p)\).

\paragraph{Flexible Acquisition with one Modality per Sample}
A direct extension is to allow acquisition of at most one additional modality per selected sample. In that case, one chooses a set
\[
\gT \subseteq \{(i,p) : i \in \{1,\dots,N\},\, p \in \gP_i^{\textnormal{miss}}\},
\]
where each pair \((i,p)\) indicates that modality \(p\) is acquired for sample \(i\). If acquisition costs are assumed uniform, one may constrain \(|\gT|=\beta\). The final score for sample \(i\) then depends on whether a pair involving \(i\) is selected:
\[
\rs_i(\gT) =
\begin{cases}
\rs_i^{\textnormal{acquired},(p)} & \text{if } (i,p) \in \gT \text{ for some } p \\
\rs_i^{\textnormal{avail}} & \text{otherwise,}
\end{cases}
\]
where \(\rs_i^{\textnormal{acquired},(p)}\) denotes the score obtained when modality \(p\) is additionally acquired for sample \(i\). The optimization problem then becomes
\[
\gT^* = \argmax_{\gT:\,|\gT|=\beta} \textnormal{Metric}(\rvy, \rvs(\gT)).
\]
Compared to the formulation in the main paper, this extension already increases the search space substantially, since one must now rank all feasible sample-modality pairs rather than only the samples.

\paragraph{Heterogeneous Acquisition Costs}
Once multiple candidate modalities are allowed, it is natural to consider heterogeneous acquisition costs. Let \(c_p\) denote the cost of acquiring modality \(p\). Then the budget constraint becomes
\[
\sum_{(i,p)\in\gT} c_p \leq \beta,
\]
or, more generally, \(\sum_{(i,p)\in\gT} c_{i,p} \leq \beta\) if acquisition costs depend on both the sample and the modality. In this case, the optimization is no longer a fixed-cardinality subset selection problem, but a knapsack-like combinatorial optimization problem in which the expected performance gain must be traded off against the cost of acquisition.

\paragraph{Acquiring Multiple Modalities per Sample}
An even more general extension allows several missing modalities to be acquired for the same sample. In this setting, one selects for each sample \(i\) a subset
\[
\gA_i \subseteq \gP_i^{\textnormal{miss}},
\]
and the resulting score becomes
\[
\rs_i(\{\gA_j\}_{j=1}^N) = f(\vx_i^{\textnormal{avail}} \cup \{\vx_i^{(p)} : p \in \gA_i\}, \vtheta).
\]
The corresponding optimization problem can be written as
\[
\{\gA_i^*\}_{i=1}^N
=
\argmax_{\{\gA_i\}_{i=1}^N}
\textnormal{Metric}\!\left(\rvy, \rvs(\{\gA_i\}_{i=1}^N)\right)
\]
subject to an appropriate budget constraint such as
\[
\sum_{i=1}^N \sum_{p \in \gA_i} c_{i,p} \leq \beta.
\]
This is substantially harder than the single-modality setting because the search space grows from subsets of samples to subsets of missing modalities for each sample, which is exponential in the number of candidate modalities.

\paragraph{Application of the Architecture to Extensions}
Conceptually, the imputation-based framework introduced in the main paper can be extended to this more general setting. For each candidate pair \((i,p)\), one may define an imputed acquired score
\[
\{\rs_{i,k}^{\textnormal{imp},(p)}\}_{k=1}^K,
\]
obtained by using \(f_{\textnormal{imp}}\) to generate plausible latent representations for missing modality \(p\), followed by the classifier \(f_C\). These imputed scores can then be used to construct modality-specific acquisition functions
\[
\textnormal{AF}(i,p),
\]
which estimate the utility of acquiring modality \(p\) for sample \(i\). In principle, the same idea could be extended to sets of modalities, although in that case interactions between modalities become important: the utility of acquiring two modalities jointly is generally not equal to the sum of their individual utilities.

Compared to the single-modality formulation studied in the main paper, flexible multi-modality acquisition introduces several additional challenges: (i) the search space expands from ranking samples to ranking sample-modality pairs or subsets of modalities, (ii) heterogeneous acquisition costs must be modeled explicitly, (iii) interactions between acquired modalities may be non-additive, and (iv) evaluating all candidate acquisitions may become computationally prohibitive. We therefore view flexible multi-modality acquisition not as a minor variant of the present problem, but as a substantial extension built on top of \ac{cama}. The single-modality setting studied in the main paper isolates the core cohort-level acquisition problem in a controlled and practically meaningful form, while providing a methodological basis for future work on more general acquisition policies.

\section{Detailed Model Performance}
\label{app:model_performance_detailed}
\begin{table}[h]
    \centering
    \small
    \caption{AUROC for detailed task-specific classes in the multi-class datasets.}
    \label{tab:auroc_multiclass_detailed}
\begin{tabular}{lcc}
        \toprule
        \textbf{Dataset} & \textbf{$M_\text{pre}$} & \textbf{$M_\text{post}$} \\
        \midrule
        Symile - Fracture & $0.5907 \pm 0.1619$ & $0.6212 \pm 0.1842$ \\
        Symile - Enl.\ Card. & $0.6241 \pm 0.1175$ & $0.6667 \pm 0.0171$ \\
        Symile - Consolid. & $0.6670 \pm 0.0416$ & $0.6862 \pm 0.0289$ \\
        Symile - Atelect. & $0.5328 \pm 0.0519$ & $0.5897 \pm 0.0149$ \\
        Symile - Edema & $0.6868 \pm 0.0313$ & $0.6993 \pm 0.0103$ \\
        Symile - Cardiom. & $0.6168 \pm 0.0443$ & $0.6863 \pm 0.0181$ \\
        Symile - LL & $0.4930 \pm 0.2566$ & $0.5218 \pm 0.1281$ \\
        Symile - LO & $0.5543 \pm 0.0685$ & $0.5702 \pm 0.0329$ \\
        Symile - Pleural Eff. & $0.5989 \pm 0.0230$ & $0.6044 \pm 0.0212$ \\
        Symile - Pneumoth. & $0.5594 \pm 0.0261$ & $0.5753 \pm 0.0227$ \\
        \addlinespace
        Haim - Fracture & $0.6251 \pm 0.1209$ & $0.6244 \pm 0.1051$ \\
        Haim - Enl.\ Card. & $0.6946 \pm 0.0347$ & $0.7090 \pm 0.0361$ \\
        Haim - Consolid. & $0.7383 \pm 0.0228$ & $0.7453 \pm 0.0183$ \\
        Haim - Atelect. & $0.6286 \pm 0.0397$ & $0.6535 \pm 0.0489$ \\
        Haim - Edema & $0.7513 \pm 0.0119$ & $0.7675 \pm 0.0113$ \\
        Haim - Cardiom. & $0.6963 \pm 0.0137$ & $0.7120 \pm 0.0144$ \\
        Haim - LL & $0.5933 \pm 0.1842$ & $0.5969 \pm 0.1829$ \\
        Haim - LO & $0.6329 \pm 0.0310$ & $0.6418 \pm 0.0390$ \\
        Haim - Pleural Eff. & $0.7082 \pm 0.0137$ & $0.7244 \pm 0.0145$ \\
        Haim - Pneumoth. & $0.7084 \pm 0.0287$ & $0.6840 \pm 0.0343$ \\
        \bottomrule
    \end{tabular}
\end{table}

\section{Runtime Analysis}
\label{app:runtime_analysis}
\begin{table*}[h]
    \centering
    \small
    \caption{Efficiency analysis for different architectures.}
    \label{tab:runtime}
    \begin{tabular}{lccc}  % @{} {}@
    \toprule
    \textbf{Architecture} & \textbf{Train (sec) $\downarrow$} & \textbf{Validation (sec) $\downarrow$} & \textbf{Parameters (M) $\downarrow$} \\
    \midrule
    \Acl{lf}                     &   0.02      &   0.015      &   86.5 \\
    \Acl{lf} w/ \acp{ddpm}       &   0.17      &   0.16       &    125 \\
    \Acl{lf} w/ \acp{bcvae}      &   0.08      &   0.08       &    313 \\
    \bottomrule
    \end{tabular}
\end{table*}

\newpage 
\section{Results for Symile with \acp{bcvae}}
\label{app:rbt_bcvae_symile}
\begin{table}[h]
    \centering
    \scriptsize
    \caption{
        Acquisition performance on Symile (\ac{auroc}) with Beta-Conditional Variational Auto Encoders. Strategies are grouped by category. Best strategy among proposed ones and baselines in bold for each column.
    }
    \label{tab:results_vae_symile_auroc_revised_sem_corr}
    \begin{tabular}{@{}l ccccc c@{}}
        \toprule
        \textbf{Strategy} & \textbf{Fracture} & \textbf{Enl. Card.} & \textbf{Consolidation} & \textbf{Atelectasis} & \textbf{Edema} & \textbf{Mean} \\
        \midrule
        \multicolumn{7}{@{}l}{\textit{Upper Bounds (for reference)}} \\
        Oracle & $2.807 \pm 0.326$ & $4.224 \pm 0.501$ & $2.716 \pm 0.147$ & $5.901 \pm 0.861$ & $2.096 \pm 0.078$ & $4.423$ \\
        True KL-Div. & $1.009 \pm 0.115$ & $0.752 \pm 0.095$ & $0.855 \pm 0.020$ & $0.689 \pm 0.092$ & $0.900 \pm 0.007$ & $0.800$ \\
        True Rank & $0.714 \pm 0.094$ & $0.728 \pm 0.070$ & $0.853 \pm 0.017$ & $0.777 \pm 0.104$ & $0.890 \pm 0.007$ & $0.719$ \\
        True Uncert. & $0.939 \pm 0.088$ & $0.735 \pm 0.113$ & $0.571 \pm 0.024$ & $0.106 \pm 0.079$ & $0.719 \pm 0.009$ & $0.555$ \\
        \addlinespace
        \midrule 
        \multicolumn{7}{@{}l}{\textit{Imputation-based (proposed)}} \\
        KL-Div. & $0.834 \pm 0.060$ & $0.420 \pm 0.109$ & $\mathbf{0.684 \pm 0.017}$ & $\mathbf{0.527 \pm 0.077}$ & $\mathbf{0.744 \pm 0.013}$ & $\mathbf{0.584}$ \\
\bottomrule
    \end{tabular}
    \vspace{1em} % Space between tabulars
    \begin{tabular}{@{}l ccccc c@{}}
        \toprule
        \textbf{Strategy} & \textbf{Cardiomegaly} & \textbf{Lung Lesion} & \textbf{Lung Opacity} & \textbf{Pneumonia} & \textbf{Pneumothorax} & \textbf{Mean} \\
        \midrule
        \multicolumn{7}{@{}l}{\textit{Upper Bounds (for reference)}} \\
        Oracle & $2.715 \pm 0.124$ & $4.677 \pm 0.799$ & $6.125 \pm 0.911$ & $4.317 \pm 0.181$ & $8.654 \pm 0.796$ & $4.423$ \\
        True KL-Div. & $0.871 \pm 0.011$ & $0.701 \pm 0.189$ & $0.582 \pm 0.146$ & $0.892 \pm 0.014$ & $0.745 \pm 0.096$ & $0.800$ \\
        True Rank & $0.843 \pm 0.015$ & $0.457 \pm 0.225$ & $0.501 \pm 0.140$ & $0.825 \pm 0.022$ & $0.603 \pm 0.102$ & $0.719$ \\
        True Uncert. & $0.701 \pm 0.016$ & $0.626 \pm 0.130$ & $0.147 \pm 0.057$ & $0.664 \pm 0.023$ & $0.343 \pm 0.036$ & $0.555$ \\
        \addlinespace
        \midrule 
        \multicolumn{7}{@{}l}{\textit{Imputation-based (proposed)}} \\
        KL-Div. & $\mathbf{0.718 \pm 0.019}$ & $0.456 \pm 0.097$ & $\mathbf{0.324 \pm 0.232}$ & $\mathbf{0.757 \pm 0.025}$ & $0.380 \pm 0.129$ & $\mathbf{0.584}$ \\
\bottomrule
    \end{tabular}
\end{table}

\begin{table}[h]
    \centering
    \scriptsize
    \caption{
        Acquisition performance on Symile (\ac{auprc}) with Beta-Conditional Variational Auto Encoders. Strategies are grouped by category. Best strategy among proposed ones and baselines in bold for each column.
    }
    \label{tab:results_vae_symile_auprc_revised_sem_corr}
    \begin{tabular}{@{}l ccccc c@{}}
        \toprule
        \textbf{Strategy} & \textbf{Fracture} & \textbf{Enl. Card.} & \textbf{Consolidation} & \textbf{Atelectasis} & \textbf{Edema} & \textbf{Mean} \\
        \midrule
        \multicolumn{7}{@{}l}{\textit{Upper Bounds (for reference)}} \\
        Oracle & $1.785 \pm 0.110$ & $3.468 \pm 0.317$ & $2.780 \pm 0.146$ & $2.632 \pm 0.146$ & $2.460 \pm 0.099$ & $4.116$ \\
        True KL-Div. & $0.798 \pm 0.050$ & $0.700 \pm 0.101$ & $0.828 \pm 0.020$ & $0.686 \pm 0.026$ & $0.888 \pm 0.007$ & $0.766$ \\
        True Rank & $0.736 \pm 0.061$ & $0.625 \pm 0.096$ & $0.822 \pm 0.022$ & $0.733 \pm 0.046$ & $0.843 \pm 0.009$ & $0.689$ \\
        True Uncert. & $0.778 \pm 0.050$ & $0.623 \pm 0.044$ & $0.606 \pm 0.034$ & $0.206 \pm 0.033$ & $0.731 \pm 0.011$ & $0.513$ \\
        \addlinespace
        \midrule 
        \multicolumn{7}{@{}l}{\textit{Imputation-based (proposed)}} \\
        KL-Div. & $0.725 \pm 0.062$ & $0.574 \pm 0.095$ & $\mathbf{0.636 \pm 0.023}$ & $\mathbf{0.599 \pm 0.034}$ & $\mathbf{0.733 \pm 0.013}$ & $\mathbf{0.624}$ \\
\bottomrule
    \end{tabular}
    \vspace{1em} % Space between tabulars
    \begin{tabular}{@{}l ccccc c@{}}
        \toprule
        \textbf{Strategy} & \textbf{Cardiomegaly} & \textbf{Lung Lesion} & \textbf{Lung Opacity} & \textbf{Pneumonia} & \textbf{Pneumothorax} & \textbf{Mean} \\
        \midrule
        \multicolumn{7}{@{}l}{\textit{Upper Bounds (for reference)}} \\
        Oracle & $2.902 \pm 0.184$ & $2.559 \pm 0.272$ & $3.156 \pm 0.152$ & $5.255 \pm 0.253$ & $14.162 \pm 1.561$ & $4.116$ \\
        True KL-Div. & $0.871 \pm 0.014$ & $0.488 \pm 0.088$ & $0.700 \pm 0.026$ & $0.879 \pm 0.021$ & $0.818 \pm 0.044$ & $0.766$ \\
        True Rank & $0.828 \pm 0.022$ & $0.533 \pm 0.068$ & $0.667 \pm 0.036$ & $0.824 \pm 0.028$ & $0.278 \pm 0.139$ & $0.689$ \\
        True Uncert. & $0.745 \pm 0.016$ & $0.183 \pm 0.045$ & $0.151 \pm 0.032$ & $0.525 \pm 0.023$ & $0.579 \pm 0.076$ & $0.513$ \\
        \addlinespace
        \midrule 
        \multicolumn{7}{@{}l}{\textit{Imputation-based (proposed)}} \\
        KL-Div. & $\mathbf{0.749 \pm 0.023}$ & $0.390 \pm 0.117$ & $\mathbf{0.584 \pm 0.023}$ & $\mathbf{0.736 \pm 0.021}$ & $0.514 \pm 0.099$ & $\mathbf{0.624}$ \\
\bottomrule
    \end{tabular}
\end{table}

\clearpage 
\section{Detailed Results for MOSEI}
\label{app:detailed_results_mosei}
% Image, Text, Audio

\begin{table}[h]
    \centering
    \scriptsize
    \caption{
        Acquisition performance on MOSEI (Image imputed by Text), showing $G_{\text{full}}$ for AUROC/AUPRC. Strategies are grouped by category. Best strategy among proposed ones and baselines in bold.
    }
    \label{tab:acquisition_strategy_performance_mosei_combined_m0m1_revised}
    \begin{tabular}{@{}l c c@{}}
        \toprule
        & \textbf{AUROC} & \textbf{AUPRC} \\
        \cmidrule(lr){2-2} \cmidrule(lr){3-3}
        \textbf{Strategy} 
        & $G_{\text{full}} \pm \text{SEM} \uparrow$ & $G_{\text{full}} \pm \text{SEM} \uparrow$ \\
        
        \midrule
        \multicolumn{3}{@{}l}{\textit{Upper Bounds (for reference)}} \\
        Oracle & $0.995 \pm 0.006$ & $0.995 \pm 0.012$ \\
        True KL-Div. & $0.777 \pm 0.005$ & $0.790 \pm 0.015$ \\
        True Rank & $0.763 \pm 0.012$ & $0.781 \pm 0.014$ \\
        True Uncert. & $0.599 \pm 0.015$ & $0.673 \pm 0.019$ \\
        \addlinespace

        \midrule 
        \multicolumn{3}{@{}l}{\textit{Imputation-based (proposed)}} \\
        KL-Div & $\mathbf{0.551 \pm 0.012}$ & $0.590 \pm 0.009$ \\
        Probability & $0.473 \pm 0.010$ & $\mathbf{0.598 \pm 0.006}$ \\
        Rank & $0.524 \pm 0.012$ & $0.560 \pm 0.012$ \\
        Uncertainty & $0.500 \pm 0.021$ & $0.567 \pm 0.009$ \\
        \addlinespace

        \midrule 
        \multicolumn{3}{@{}l}{\textit{Baselines (no imputation)}} \\
        Uncertainty & $0.507 \pm 0.015$ & $0.565 \pm 0.018$ \\
        Probability & $0.451 \pm 0.013$ & $0.578 \pm 0.009$ \\
        Random & $0.521 \pm 0.006$ & $0.575 \pm 0.008$ \\
        \bottomrule
    \end{tabular}
\end{table}

\begin{table}[h]
    \centering
    \scriptsize
    \caption{
        Acquisition performance on MOSEI (Image imputed by Audio), showing $G_{\text{full}}$ for AUROC/AUPRC. Strategies are grouped by category. Best strategy among proposed ones and baselines in bold.
    }
    \label{tab:acquisition_strategy_performance_mosei_combined_m0m2_revised}
    \begin{tabular}{@{}l c c@{}}
        \toprule
        & \textbf{AUROC} & \textbf{AUPRC} \\
        \cmidrule(lr){2-2} \cmidrule(lr){3-3}
        \textbf{Strategy} 
        & $G_{\text{full}} \pm \text{SEM} \uparrow$ & $G_{\text{full}} \pm \text{SEM} \uparrow$ \\
        
        \midrule
        \multicolumn{3}{@{}l}{\textit{Upper Bounds (for reference)}} \\
        Oracle & $1.052 \pm 0.007$ & $1.011 \pm 0.010$ \\
        True KL-Div. & $0.785 \pm 0.009$ & $0.803 \pm 0.005$ \\
        True Rank & $0.783 \pm 0.010$ & $0.802 \pm 0.012$ \\
        True Uncert. & $0.672 \pm 0.007$ & $0.742 \pm 0.006$ \\
        \addlinespace

        \midrule 
        \multicolumn{3}{@{}l}{\textit{Imputation-based (proposed)}} \\
        KL-Div & $0.566 \pm 0.011$ & $0.601 \pm 0.009$ \\
        Probability & $0.547 \pm 0.002$ & $\mathbf{0.629 \pm 0.004}$ \\
        Rank & $\mathbf{0.576 \pm 0.013}$ & $0.614 \pm 0.007$ \\
        Uncertainty & $0.545 \pm 0.009$ & $0.586 \pm 0.007$ \\
        \addlinespace

        \midrule 
        \multicolumn{3}{@{}l}{\textit{Baselines (no imputation)}} \\
        Uncertainty & $0.545 \pm 0.009$ & $0.601 \pm 0.014$ \\
        Probability & $0.526 \pm 0.014$ & $0.616 \pm 0.008$ \\
        Random & $0.553 \pm 0.008$ & $0.603 \pm 0.005$ \\
        \bottomrule
    \end{tabular}
\end{table}

\begin{table}[h]
    \centering
    \scriptsize
    \caption{
        Acquisition performance on MOSEI (Image imputed by Text and Audio), showing $G_{\text{full}}$ for AUROC/AUPRC. Strategies are grouped by category. Best strategy among proposed ones and baselines in bold.
    }
    \label{tab:acquisition_strategy_performance_mosei_combined_m0m1m2_revised}
    \begin{tabular}{@{}l c c@{}}
        \toprule
        & \textbf{AUROC} & \textbf{AUPRC} \\
        \cmidrule(lr){2-2} \cmidrule(lr){3-3}
        \textbf{Strategy} 
        & $G_{\text{full}} \pm \text{SEM} \uparrow$ & $G_{\text{full}} \pm \text{SEM} \uparrow$ \\
        
        \midrule
        \multicolumn{3}{@{}l}{\textit{Upper Bounds (for reference)}} \\
        Oracle & $1.321 \pm 0.009$ & $1.315 \pm 0.012$ \\
        True KL-Div. & $0.979 \pm 0.006$ & $0.862 \pm 0.006$ \\
        True Rank & $0.960 \pm 0.005$ & $0.811 \pm 0.006$ \\
        True Uncert. & $0.716 \pm 0.004$ & $0.750 \pm 0.004$ \\
        \addlinespace

        \midrule 
        \multicolumn{3}{@{}l}{\textit{Imputation-based (proposed)}} \\
        KL-Div & $\mathbf{0.603 \pm 0.007}$ & $0.627 \pm 0.009$ \\
        Probability & $0.513 \pm 0.002$ & $0.610 \pm 0.003$ \\
        Rank & $0.463 \pm 0.005$ & $0.491 \pm 0.005$ \\
        Uncertainty & $0.499 \pm 0.004$ & $0.452 \pm 0.004$ \\
        \addlinespace

        \midrule 
        \multicolumn{3}{@{}l}{\textit{Baselines (no imputation)}} \\
        Uncertainty & $0.513 \pm 0.004$ & $0.480 \pm 0.005$ \\
        Probability & $0.534 \pm 0.008$ & $\mathbf{0.676 \pm 0.007}$ \\
        Random & $0.489 \pm 0.002$ & $0.527 \pm 0.002$ \\
        \bottomrule
    \end{tabular}
\end{table}

% Text by Image
\begin{table}[h]
    \centering
    \scriptsize
    \caption{
        Acquisition performance on MOSEI (Text imputed by Image), showing $G_{\text{full}}$ for AUROC/AUPRC. Strategies are grouped by category. Best strategy among proposed ones and baselines in bold.
    }
    \label{tab:acquisition_strategy_performance_mosei_combined_m1m0_revised}
    \begin{tabular}{@{}l c c@{}}
        \toprule
        & \textbf{AUROC} & \textbf{AUPRC} \\
        \cmidrule(lr){2-2} \cmidrule(lr){3-3}
        \textbf{Strategy} 
        & $G_{\text{full}} \pm \text{SEM} \uparrow$ & $G_{\text{full}} \pm \text{SEM} \uparrow$ \\
        
        \midrule
        \multicolumn{3}{@{}l}{\textit{Upper Bounds (for reference)}} \\
        Oracle & $1.613 \pm 0.269$ & $1.493 \pm 0.166$ \\
        True KL-Div. & $0.845 \pm 0.018$ & $0.843 \pm 0.010$ \\
        True Rank & $0.772 \pm 0.024$ & $0.784 \pm 0.025$ \\
        True Uncert. & $0.633 \pm 0.060$ & $0.697 \pm 0.030$ \\
        \addlinespace

        \midrule 
        \multicolumn{3}{@{}l}{\textit{Imputation-based (proposed)}} \\
        KL-Div & $\mathbf{0.900 \pm 0.023}$ & $\mathbf{0.896 \pm 0.022}$ \\
        Probability & $0.806 \pm 0.022$ & $0.861 \pm 0.023$ \\
        Rank & $0.309 \pm 0.089$ & $0.420 \pm 0.065$ \\
        Uncertainty & $0.651 \pm 0.049$ & $0.747 \pm 0.030$ \\
        \addlinespace

        \midrule 
        \multicolumn{3}{@{}l}{\textit{Baselines (no imputation)}} \\
        Uncertainty & $0.466 \pm 0.045$ & $0.537 \pm 0.014$ \\
        Probability & $0.418 \pm 0.055$ & $0.532 \pm 0.045$ \\
        Random & $0.417 \pm 0.061$ & $0.489 \pm 0.051$ \\
        \bottomrule
    \end{tabular}
\end{table}

% Text by Audio
\begin{table}[h]
    \centering
    \scriptsize
    \caption{
        Acquisition performance on MOSEI (Text imputed by Audio), showing $G_{\text{full}}$ for AUROC/AUPRC. Strategies are grouped by category. Best strategy among proposed ones and baselines in bold.
    }
    \label{tab:acquisition_strategy_performance_mosei_combined_m1m2_revised}
    \begin{tabular}{@{}l c c@{}}
        \toprule
        & \textbf{AUROC} & \textbf{AUPRC} \\
        \cmidrule(lr){2-2} \cmidrule(lr){3-3}
        \textbf{Strategy} 
        & $G_{\text{full}} \pm \text{SEM} \uparrow$ & $G_{\text{full}} \pm \text{SEM} \uparrow$ \\
        
        \midrule
        \multicolumn{3}{@{}l}{\textit{Upper Bounds (for reference)}} \\
        Oracle & $8.867 \pm 1.712$ & $16.066 \pm 2.592$ \\
        True KL-Div. & $0.645 \pm 0.051$ & $0.387 \pm 0.136$ \\
        True Rank & $0.494 \pm 0.147$ & $0.015 \pm 0.262$ \\
        True Uncert. & $0.913 \pm 0.121$ & $0.893 \pm 0.141$ \\
        \addlinespace

        \midrule 
        \multicolumn{3}{@{}l}{\textit{Imputation-based (proposed)}} \\
        KL-Div & $2.962 \pm 0.925$ & $4.582 \pm 1.938$ \\
        Probability & $\mathbf{3.157 \pm 1.229}$ & $\mathbf{6.861 \pm 1.806}$ \\
        Rank & $-0.413 \pm 0.421$ & $-1.134 \pm 0.463$ \\
        Uncertainty & $1.592 \pm 0.350$ & $3.844 \pm 0.806$ \\
        \addlinespace

        \midrule 
        \multicolumn{3}{@{}l}{\textit{Baselines (no imputation)}} \\
        Uncertainty & $0.591 \pm 0.147$ & $0.473 \pm 0.204$ \\
        Probability & $0.662 \pm 0.071$ & $0.971 \pm 0.014$ \\
        Random & $0.316 \pm 0.059$ & $0.376 \pm 0.086$ \\
        \bottomrule
    \end{tabular}
\end{table}

% Text by Image and Audio
\begin{table}[h]
    \centering
    \scriptsize
    \caption{
        Acquisition performance on MOSEI (Text imputed by Image and Audio), showing $G_{\text{full}}$ for AUROC/AUPRC. Strategies are grouped by category. Best strategy among proposed ones and baselines in bold.
    }
    \label{tab:acquisition_strategy_performance_mosei_combined_m1m0m2_revised}
    \begin{tabular}{@{}l c c@{}}
        \toprule
        & \textbf{AUROC} & \textbf{AUPRC} \\
        \cmidrule(lr){2-2} \cmidrule(lr){3-3}
        \textbf{Strategy} 
        & $G_{\text{full}} \pm \text{SEM} \uparrow$ & $G_{\text{full}} \pm \text{SEM} \uparrow$ \\
        
        \midrule
        \multicolumn{3}{@{}l}{\textit{Upper Bounds (for reference)}} \\
        Oracle & $1.207 \pm 0.012$ & $1.280 \pm 0.013$ \\
        True KL-Div. & $0.851 \pm 0.002$ & $0.842 \pm 0.004$ \\
        True Rank & $0.836 \pm 0.004$ & $0.840 \pm 0.005$ \\
        True Uncert. & $0.649 \pm 0.009$ & $0.691 \pm 0.009$ \\
        \addlinespace

        \midrule 
        \multicolumn{3}{@{}l}{\textit{Imputation-based (proposed)}} \\
        KL-Div & $\mathbf{0.892 \pm 0.002}$ & $\mathbf{0.894 \pm 0.004}$ \\
        Probability & $0.665 \pm 0.004$ & $0.727 \pm 0.003$ \\
        Rank & $0.489 \pm 0.003$ & $0.520 \pm 0.003$ \\
        Uncertainty & $0.662 \pm 0.008$ & $0.720 \pm 0.007$ \\
        \addlinespace

        \midrule 
        \multicolumn{3}{@{}l}{\textit{Baselines (no imputation)}} \\
        Uncertainty & $0.538 \pm 0.009$ & $0.567 \pm 0.009$ \\
        Probability & $0.379 \pm 0.003$ & $0.462 \pm 0.004$ \\
        Random & $0.512 \pm 0.002$ & $0.531 \pm 0.003$ \\
        \bottomrule
    \end{tabular}
\end{table}

% Audio by Image
\begin{table}[h]
    \centering
    \scriptsize
    \caption{
        Acquisition performance on MOSEI (Audio imputed by Image), showing $G_{\text{full}}$ for AUROC/AUPRC. Strategies are grouped by category. Best strategy among proposed ones and baselines in bold.
    }
    \label{tab:acquisition_strategy_performance_mosei_combined_m2m0_revised}
    \begin{tabular}{@{}l c c@{}}
        \toprule
        & \textbf{AUROC} & \textbf{AUPRC} \\
        \cmidrule(lr){2-2} \cmidrule(lr){3-3}
        \textbf{Strategy} 
        & $G_{\text{full}} \pm \text{SEM} \uparrow$ & $G_{\text{full}} \pm \text{SEM} \uparrow$ \\
        
        \midrule
        \multicolumn{3}{@{}l}{\textit{Upper Bounds (for reference)}} \\
        Oracle & $1.238 \pm 0.124$ & $1.207 \pm 0.093$ \\
        True KL-Div. & $0.826 \pm 0.014$ & $0.821 \pm 0.019$ \\
        True Rank & $0.752 \pm 0.030$ & $0.780 \pm 0.025$ \\
        True Uncert. & $0.544 \pm 0.034$ & $0.627 \pm 0.023$ \\
        \addlinespace

        \midrule 
        \multicolumn{3}{@{}l}{\textit{Imputation-based (proposed)}} \\
        KL-Div & $\mathbf{0.800 \pm 0.012}$ & $\mathbf{0.803 \pm 0.017}$ \\
        Probability & $0.684 \pm 0.020$ & $0.737 \pm 0.024$ \\
        Rank & $0.326 \pm 0.072$ & $0.445 \pm 0.052$ \\
        Uncertainty & $0.552 \pm 0.035$ & $0.625 \pm 0.023$ \\
        \addlinespace

        \midrule 
        \multicolumn{3}{@{}l}{\textit{Baselines (no imputation)}} \\
        Uncertainty & $0.436 \pm 0.043$ & $0.502 \pm 0.016$ \\
        Probability & $0.374 \pm 0.034$ & $0.510 \pm 0.034$ \\
        Random & $0.397 \pm 0.052$ & $0.478 \pm 0.039$ \\
        \bottomrule
    \end{tabular}
\end{table}

% Audio by Text
\begin{table}[h]
    \centering
    \scriptsize
    \caption{
        Acquisition performance on MOSEI (Audio imputed by Text), showing $G_{\text{full}}$ for AUROC/AUPRC. Strategies are grouped by category. Best strategy among proposed ones and baselines in bold.
    }
    \label{tab:acquisition_strategy_performance_mosei_combined_m2m1_revised}
    \begin{tabular}{@{}l c c@{}}
        \toprule
        & \textbf{AUROC} & \textbf{AUPRC} \\
        \cmidrule(lr){2-2} \cmidrule(lr){3-3}
        \textbf{Strategy} 
        & $G_{\text{full}} \pm \text{SEM} \uparrow$ & $G_{\text{full}} \pm \text{SEM} \uparrow$ \\
        
        \midrule
        \multicolumn{3}{@{}l}{\textit{Upper Bounds (for reference)}} \\
        Oracle & $4.955 \pm 0.417$ & $6.275 \pm 0.948$ \\
        True KL-Div. & $0.815 \pm 0.043$ & $0.817 \pm 0.057$ \\
        True Rank & $0.563 \pm 0.073$ & $0.645 \pm 0.098$ \\
        True Uncert. & $0.520 \pm 0.026$ & $0.604 \pm 0.059$ \\
        \addlinespace

        \midrule 
        \multicolumn{3}{@{}l}{\textit{Imputation-based (proposed)}} \\
        KL-Div & $2.305 \pm 0.360$ & $2.335 \pm 0.499$ \\
        Probability & $\mathbf{2.306 \pm 0.259}$ & $\mathbf{2.571 \pm 0.483}$ \\
        Rank & $-0.210 \pm 0.123$ & $-0.093 \pm 0.135$ \\
        Uncertainty & $1.154 \pm 0.141$ & $1.626 \pm 0.303$ \\
        \addlinespace

        \midrule 
        \multicolumn{3}{@{}l}{\textit{Baselines (no imputation)}} \\
        Uncertainty & $0.519 \pm 0.023$ & $0.604 \pm 0.044$ \\
        Probability & $0.416 \pm 0.007$ & $0.511 \pm 0.018$ \\
        Random & $0.326 \pm 0.024$ & $0.439 \pm 0.023$ \\
        \bottomrule
    \end{tabular}
\end{table}

% Audio by Image and Text
\begin{table}[h]
    \centering
    \scriptsize
    \caption{
        Acquisition performance on MOSEI (Audio imputed by Image and Text), showing $G_{\text{full}}$ for AUROC/AUPRC. Strategies are grouped by category. Best strategy among proposed ones and baselines in bold.
    }
    \label{tab:acquisition_strategy_performance_mosei_combined_m2m0m1_revised}
    \begin{tabular}{@{}l c c@{}}
        \toprule
        & \textbf{AUROC} & \textbf{AUPRC} \\
        \cmidrule(lr){2-2} \cmidrule(lr){3-3}
        \textbf{Strategy} 
        & $G_{\text{full}} \pm \text{SEM} \uparrow$ & $G_{\text{full}} \pm \text{SEM} \uparrow$ \\
        
        \midrule
        \multicolumn{3}{@{}l}{\textit{Upper Bounds (for reference)}} \\
        Oracle & $1.215 \pm 0.010$ & $1.275 \pm 0.011$ \\
        True KL-Div. & $0.865 \pm 0.002$ & $0.850 \pm 0.004$ \\
        True Rank & $0.833 \pm 0.004$ & $0.837 \pm 0.005$ \\
        True Uncert. & $0.645 \pm 0.007$ & $0.693 \pm 0.006$ \\
        \addlinespace

        \midrule 
        \multicolumn{3}{@{}l}{\textit{Imputation-based (proposed)}} \\
        KL-Div & $\mathbf{0.857 \pm 0.002}$ & $\mathbf{0.846 \pm 0.004}$ \\
        Probability & $0.667 \pm 0.003$ & $0.725 \pm 0.003$ \\
        Rank & $0.459 \pm 0.004$ & $0.489 \pm 0.004$ \\
        Uncertainty & $0.646 \pm 0.007$ & $0.694 \pm 0.007$ \\
        \addlinespace

        \midrule 
        \multicolumn{3}{@{}l}{\textit{Baselines (no imputation)}} \\
        Uncertainty & $0.536 \pm 0.008$ & $0.566 \pm 0.008$ \\
        Probability & $0.368 \pm 0.005$ & $0.462 \pm 0.005$ \\
        Random & $0.503 \pm 0.003$ & $0.530 \pm 0.003$ \\
        \bottomrule
    \end{tabular}
\end{table}

\clearpage 
\section{Detailed Results for MIMIC Symile}
\label{app:detailed_results_mimic_symile}
% Imaage, Laab EECG
\begin{table}[h]
    \centering
    \scriptsize
    \caption{
        Acquisition performance on MIMIC Symile for AUROC, showing $G_{\text{full}}$. Strategies are grouped by category. Best strategy among proposed and baseline methods in bold for each column.
    }
    \label{tab:acquisition_strategy_performance_auroc_mimic_symile_revised}
    \begin{tabular}{@{}l ccccc c@{}}
        \toprule
        \textbf{Strategy} & \textbf{Fracture} & \textbf{Enl. Card.} & \textbf{Consolidation} & \textbf{Atelectasis} & \textbf{Edema} & \textbf{Mean} \\
        \midrule
        \multicolumn{7}{@{}l}{\textit{Upper Bounds (for reference)}} \\
        Oracle & $2.876 \pm 0.368$ & $3.722 \pm 0.370$ & $2.856 \pm 0.147$ & $4.862 \pm 0.356$ & $2.793 \pm 0.324$ & $4.580$ \\
        True KL-Div. & $1.029 \pm 0.165$ & $1.019 \pm 0.051$ & $0.885 \pm 0.025$ & $0.841 \pm 0.045$ & $0.920 \pm 0.011$ & $0.883$ \\
        True Rank & $0.963 \pm 0.153$ & $0.924 \pm 0.058$ & $0.946 \pm 0.025$ & $0.802 \pm 0.068$ & $0.887 \pm 0.022$ & $0.811$ \\
        True Uncert. & $0.915 \pm 0.138$ & $0.763 \pm 0.052$ & $0.749 \pm 0.031$ & $0.152 \pm 0.052$ & $0.648 \pm 0.012$ & $0.481$ \\
        \addlinespace
        \midrule 
        \multicolumn{7}{@{}l}{\textit{Imputation-based (proposed)}} \\
        KL-Div & $0.838 \pm 0.164$ & $\mathbf{0.882 \pm 0.064}$ & $\mathbf{0.706 \pm 0.021}$ & $\mathbf{0.779 \pm 0.114}$ & $\mathbf{0.893 \pm 0.080}$ & $\mathbf{0.833}$ \\
\bottomrule
    \end{tabular}
    \vspace{1em} % Space between tabulars
    \begin{tabular}{@{}l ccccc c@{}}
        \toprule
        \textbf{Strategy} & \textbf{Cardiomegaly} & \textbf{Lung Lesion} & \textbf{Lung Opacity} & \textbf{Pneumonia} & \textbf{Pneumothorax} & \textbf{Mean} \\
        \midrule
        \multicolumn{7}{@{}l}{\textit{Upper Bounds (for reference)}} \\
        Oracle & $2.787 \pm 0.139$ & $4.974 \pm 0.581$ & $6.657 \pm 0.649$ & $4.817 \pm 0.430$ & $9.461 \pm 1.049$ & $4.580$ \\
        True KL-Div. & $0.885 \pm 0.011$ & $0.753 \pm 0.179$ & $0.750 \pm 0.087$ & $0.837 \pm 0.043$ & $0.910 \pm 0.054$ & $0.883$ \\
        True Rank & $0.878 \pm 0.019$ & $0.411 \pm 0.216$ & $0.793 \pm 0.056$ & $0.902 \pm 0.029$ & $0.605 \pm 0.053$ & $0.811$ \\
        True Uncert. & $0.524 \pm 0.025$ & $0.251 \pm 0.177$ & $0.212 \pm 0.054$ & $0.728 \pm 0.023$ & $-0.136 \pm 0.065$ & $0.481$ \\
        \addlinespace
        \midrule 
        \multicolumn{7}{@{}l}{\textit{Imputation-based (proposed)}} \\
        KL-Div & $\mathbf{0.747 \pm 0.039}$ & $\mathbf{1.266 \pm 0.258}$ & $\mathbf{0.683 \pm 0.106}$ & $\mathbf{0.761 \pm 0.060}$ & $0.773 \pm 0.134$ & $\mathbf{0.833}$ \\
\bottomrule
    \end{tabular}
\end{table}

\begin{table}[h]
    \centering
    \scriptsize
    \caption{
        Acquisition performance on MIMIC Symile for AUPRC, showing $G_{\text{full}}$. Strategies are grouped by category. Best strategy among proposed and baseline methods in bold for each column.
    }
    \label{tab:acquisition_strategy_performance_auprc_mimic_symile_revised}
    \begin{tabular}{@{}l ccccc c@{}}
        \toprule
        \textbf{Strategy} & \textbf{Fracture} & \textbf{Enl. Card.} & \textbf{Consolidation} & \textbf{Atelectasis} & \textbf{Edema} & \textbf{Mean} \\
        \midrule
        \multicolumn{7}{@{}l}{\textit{Upper Bounds (for reference)}} \\
        Oracle & $2.579 \pm 0.343$ & $2.964 \pm 0.201$ & $4.784 \pm 1.130$ & $3.093 \pm 0.201$ & $3.015 \pm 0.141$ & $4.231$ \\
        True KL-Div. & $0.883 \pm 0.094$ & $0.970 \pm 0.038$ & $0.933 \pm 0.057$ & $0.781 \pm 0.038$ & $0.939 \pm 0.013$ & $0.871$ \\
        True Rank & $0.659 \pm 0.176$ & $0.858 \pm 0.033$ & $0.903 \pm 0.090$ & $0.720 \pm 0.064$ & $0.897 \pm 0.022$ & $0.776$ \\
        True Uncert. & $0.604 \pm 0.073$ & $0.812 \pm 0.044$ & $0.709 \pm 0.036$ & $0.050 \pm 0.039$ & $0.645 \pm 0.015$ & $0.450$ \\
        \addlinespace
        \midrule 
        \multicolumn{7}{@{}l}{\textit{Imputation-based (proposed)}} \\
        KL-Div & $\mathbf{0.770 \pm 0.210}$ & $\mathbf{0.899 \pm 0.044}$ & $\mathbf{0.895 \pm 0.159}$ & $\mathbf{0.684 \pm 0.063}$ & $\mathbf{0.832 \pm 0.039}$ & $\mathbf{0.777}$ \\
\bottomrule
    \end{tabular}
    \vspace{1em} % Space between tabulars
    \begin{tabular}{@{}l ccccc c@{}}
        \toprule
        \textbf{Strategy} & \textbf{Cardiomegaly} & \textbf{Lung Lesion} & \textbf{Lung Opacity} & \textbf{Pneumonia} & \textbf{Pneumothorax} & \textbf{Mean} \\
        \midrule
        \multicolumn{7}{@{}l}{\textit{Upper Bounds (for reference)}} \\
        Oracle & $2.746 \pm 0.110$ & $2.520 \pm 0.250$ & $4.092 \pm 0.259$ & $5.895 \pm 0.406$ & $10.623 \pm 0.708$ & $4.231$ \\
        True KL-Div. & $0.853 \pm 0.010$ & $0.828 \pm 0.073$ & $0.792 \pm 0.037$ & $0.906 \pm 0.032$ & $0.827 \pm 0.043$ & $0.871$ \\
        True Rank & $0.882 \pm 0.018$ & $0.676 \pm 0.088$ & $0.771 \pm 0.045$ & $0.911 \pm 0.030$ & $0.483 \pm 0.075$ & $0.776$ \\
        True Uncert. & $0.473 \pm 0.029$ & $0.181 \pm 0.067$ & $0.140 \pm 0.029$ & $0.595 \pm 0.024$ & $0.293 \pm 0.052$ & $0.450$ \\
        \addlinespace
        \midrule 
        \multicolumn{7}{@{}l}{\textit{Imputation-based (proposed)}} \\
        KL-Div & $\mathbf{0.729 \pm 0.034}$ & $\mathbf{0.896 \pm 0.146}$ & $\mathbf{0.722 \pm 0.043}$ & $\mathbf{0.763 \pm 0.059}$ & $0.581 \pm 0.084$ & $\mathbf{0.777}$ \\
\bottomrule
    \end{tabular}
\end{table}

\begin{table}[h]
    \centering
    \scriptsize
    \caption{
        Acquisition performance on MIMIC Symile for AUROC (Image imputed by Lab), showing $G_{\text{full}}$. Strategies are grouped by category. Best strategy among proposed and baseline methods in bold for each column.
    }
    \label{tab:acquisition_strategy_performance_auroc_mimic_haim_m0m1_revised_v2}
    \begin{tabular}{@{}l ccccc c@{}}
        \toprule
        \textbf{Strategy} & \textbf{Fracture} & \textbf{Enl. Card.} & \textbf{Consolidation} & \textbf{Atelectasis} & \textbf{Edema} & \textbf{Mean} \\
        \midrule
        \multicolumn{7}{@{}l}{\textit{Upper Bounds (for reference)}} \\
        Oracle & $4.144 \pm 2.172$ & $2.120 \pm 0.465$ & $2.803 \pm 0.306$ & $2.381 \pm 0.215$ & $1.684 \pm 0.070$ & $4.104$ \\
        True KL-Div. & $0.728 \pm 0.137$ & $0.816 \pm 0.011$ & $0.611 \pm 0.084$ & $0.837 \pm 0.028$ & $0.813 \pm 0.009$ & $0.825$ \\
        True Rank & $1.062 \pm 0.643$ & $0.614 \pm 0.111$ & $0.591 \pm 0.157$ & $0.816 \pm 0.065$ & $0.765 \pm 0.024$ & $0.707$ \\
        True Uncert. & $0.686 \pm 0.374$ & $0.606 \pm 0.068$ & $0.637 \pm 0.140$ & $0.496 \pm 0.148$ & $0.619 \pm 0.024$ & $0.601$ \\
        \addlinespace
        \midrule 
        \multicolumn{7}{@{}l}{\textit{Imputation-based (proposed)}} \\
        KL-Div & $-1.410 \pm 0.037$ & $0.477 \pm 0.108$ & $0.282 \pm 0.146$ & $\mathbf{0.626 \pm 0.137}$ & $0.503 \pm 0.022$ & $0.247$ \\
\bottomrule
    \end{tabular}
    \vspace{1em} % Space between tabulars
    \begin{tabular}{@{}l ccccc c@{}}
        \toprule
        \textbf{Strategy} & \textbf{Cardiomegaly} & \textbf{Lung Lesion} & \textbf{Lung Opacity} & \textbf{Pneumonia} & \textbf{Pneumothorax} & \textbf{Mean} \\
        \midrule
        \multicolumn{7}{@{}l}{\textit{Upper Bounds (for reference)}} \\
        Oracle & $1.793 \pm 0.077$ & $5.417 \pm 4.001$ & $7.073 \pm 2.202$ & $4.743 \pm 1.268$ & $8.881 \pm 1.894$ & $4.104$ \\
        True KL-Div. & $0.822 \pm 0.010$ & $1.543 \pm 0.689$ & $0.663 \pm 0.095$ & $0.884 \pm 0.063$ & $0.530 \pm 0.109$ & $0.825$ \\
        True Rank & $0.663 \pm 0.042$ & $1.291 \pm 0.557$ & $0.530 \pm 0.154$ & $0.705 \pm 0.170$ & $0.035 \pm 0.068$ & $0.707$ \\
        True Uncert. & $0.272 \pm 0.091$ & $1.128 \pm 0.870$ & $0.622 \pm 0.187$ & $0.687 \pm 0.038$ & $0.254 \pm 0.213$ & $0.601$ \\
        \addlinespace
        \midrule 
        \multicolumn{7}{@{}l}{\textit{Imputation-based (proposed)}} \\
        KL-Div & $\mathbf{0.492 \pm 0.030}$ & $\mathbf{0.847 \pm 0.033}$ & $0.459 \pm 0.146$ & $\mathbf{0.481 \pm 0.158}$ & $-0.285 \pm 0.221$ & $0.247$ \\
\bottomrule
    \end{tabular}
\end{table}

\begin{table}[h]
    \centering
    \scriptsize
    \caption{
        Acquisition performance on MIMIC Symile for AUPRC (Image imputed by Lab), showing $G_{\text{full}}$. Strategies are grouped by category. Best strategy among proposed and baseline methods in bold for each column.
    }
    \label{tab:acquisition_strategy_performance_auprc_mimic_haim_m0m1_revised_v2}
    \begin{tabular}{@{}l ccccc c@{}}
        \toprule
        \textbf{Strategy} & \textbf{Fracture} & \textbf{Enl. Card.} & \textbf{Consolidation} & \textbf{Atelectasis} & \textbf{Edema} & \textbf{Mean} \\
        \midrule
        \multicolumn{7}{@{}l}{\textit{Upper Bounds (for reference)}} \\
        Oracle & $3.844 \pm 2.595$ & $1.792 \pm 0.288$ & $2.826 \pm 0.677$ & $1.930 \pm 0.221$ & $1.778 \pm 0.101$ & $4.862$ \\
        True KL-Div. & $0.869 \pm 0.023$ & $0.827 \pm 0.021$ & $0.620 \pm 0.156$ & $0.812 \pm 0.024$ & $0.775 \pm 0.013$ & $0.770$ \\
        True Rank & $1.398 \pm 0.874$ & $0.566 \pm 0.140$ & $0.492 \pm 0.345$ & $0.827 \pm 0.034$ & $0.759 \pm 0.024$ & $0.575$ \\
        True Uncert. & $0.712 \pm 0.210$ & $0.650 \pm 0.111$ & $0.655 \pm 0.118$ & $0.538 \pm 0.165$ & $0.579 \pm 0.041$ & $0.498$ \\
        \addlinespace
        \midrule 
        \multicolumn{7}{@{}l}{\textit{Imputation-based (proposed)}} \\
        KL-Div & $-1.450 \pm 1.121$ & $0.464 \pm 0.132$ & $0.149 \pm 0.370$ & $\mathbf{0.652 \pm 0.149}$ & $0.610 \pm 0.010$ & $0.131$ \\
\bottomrule
    \end{tabular}
    \vspace{1em} % Space between tabulars
    \begin{tabular}{@{}l ccccc c@{}}
        \toprule
        \textbf{Strategy} & \textbf{Cardiomegaly} & \textbf{Lung Lesion} & \textbf{Lung Opacity} & \textbf{Pneumonia} & \textbf{Pneumothorax} & \textbf{Mean} \\
        \midrule
        \multicolumn{7}{@{}l}{\textit{Upper Bounds (for reference)}} \\
        Oracle & $1.749 \pm 0.092$ & $2.456 \pm 1.140$ & $7.505 \pm 3.891$ & $4.470 \pm 1.249$ & $20.270 \pm 6.707$ & $4.862$ \\
        True KL-Div. & $0.774 \pm 0.017$ & $1.035 \pm 0.214$ & $0.413 \pm 0.176$ & $0.793 \pm 0.046$ & $0.781 \pm 0.044$ & $0.770$ \\
        True Rank & $0.667 \pm 0.037$ & $0.888 \pm 0.274$ & $0.326 \pm 0.248$ & $0.669 \pm 0.058$ & $-0.837 \pm 0.865$ & $0.575$ \\
        True Uncert. & $0.269 \pm 0.134$ & $0.598 \pm 0.430$ & $0.291 \pm 0.099$ & $0.562 \pm 0.027$ & $0.129 \pm 0.407$ & $0.498$ \\
        \addlinespace
        \midrule 
        \multicolumn{7}{@{}l}{\textit{Imputation-based (proposed)}} \\
        KL-Div & $\mathbf{0.507 \pm 0.059}$ & $\mathbf{0.742 \pm 0.101}$ & $0.382 \pm 0.251$ & $0.639 \pm 0.066$ & $-1.388 \pm 1.198$ & $0.131$ \\
\bottomrule
    \end{tabular}
\end{table}

\begin{table}[h]
    \centering
    \scriptsize
    \caption{
        Acquisition performance on MIMIC Symile for AUROC (Image imputed by ECG), showing $G_{\text{full}}$. Strategies are grouped by category. Best strategy among proposed and baseline methods in bold for each column.
    }
    \label{tab:acquisition_strategy_performance_auroc_mimic_haim_m0m2_revised_v2}
    \begin{tabular}{@{}l ccccc c@{}}
        \toprule
        \textbf{Strategy} & \textbf{Fracture} & \textbf{Enl. Card.} & \textbf{Consolidation} & \textbf{Atelectasis} & \textbf{Edema} & \textbf{Mean} \\
        \midrule
        \multicolumn{7}{@{}l}{\textit{Upper Bounds (for reference)}} \\
        Oracle & $4.028 \pm 2.600$ & $1.959 \pm 0.358$ & $1.960 \pm 0.164$ & $3.526 \pm 0.947$ & $1.639 \pm 0.050$ & $3.892$ \\
        True KL-Div. & $0.963 \pm 0.224$ & $0.852 \pm 0.051$ & $0.725 \pm 0.030$ & $0.614 \pm 0.152$ & $0.777 \pm 0.003$ & $0.874$ \\
        True Rank & $0.192 \pm 0.075$ & $0.559 \pm 0.101$ & $0.753 \pm 0.013$ & $-0.784 \pm 1.091$ & $0.719 \pm 0.043$ & $0.327$ \\
        True Uncert. & $0.934 \pm 0.189$ & $0.700 \pm 0.131$ & $0.717 \pm 0.039$ & $-0.853 \pm 0.584$ & $0.658 \pm 0.050$ & $0.024$ \\
        \addlinespace
        \midrule 
        \multicolumn{7}{@{}l}{\textit{Imputation-based (proposed)}} \\
        KL-Div & $0.425 \pm 0.323$ & $\mathbf{0.561 \pm 0.079}$ & $\mathbf{0.769 \pm 0.096}$ & $-0.353 \pm 0.827$ & $\mathbf{0.621 \pm 0.030}$ & $\mathbf{0.669}$ \\
\bottomrule
    \end{tabular}
    \vspace{1em} % Space between tabulars
    \begin{tabular}{@{}l ccccc c@{}}
        \toprule
        \textbf{Strategy} & \textbf{Cardiomegaly} & \textbf{Lung Lesion} & \textbf{Lung Opacity} & \textbf{Pneumonia} & \textbf{Pneumothorax} & \textbf{Mean} \\
        \midrule
        \multicolumn{7}{@{}l}{\textit{Upper Bounds (for reference)}} \\
        Oracle & $2.263 \pm 0.151$ & $9.495 \pm 7.441$ & $4.112 \pm 0.961$ & $2.196 \pm 0.132$ & $7.745 \pm 2.386$ & $3.892$ \\
        True KL-Div. & $0.777 \pm 0.016$ & $1.758 \pm 1.206$ & $0.881 \pm 0.085$ & $0.725 \pm 0.019$ & $0.664 \pm 0.088$ & $0.874$ \\
        True Rank & $0.560 \pm 0.050$ & $-0.731 \pm 0.850$ & $0.808 \pm 0.103$ & $0.777 \pm 0.022$ & $0.416 \pm 0.209$ & $0.327$ \\
        True Uncert. & $0.741 \pm 0.021$ & $-3.450 \pm 3.932$ & $-0.150 \pm 0.160$ & $0.727 \pm 0.010$ & $0.213 \pm 0.412$ & $0.024$ \\
        \addlinespace
        \midrule 
        \multicolumn{7}{@{}l}{\textit{Imputation-based (proposed)}} \\
        KL-Div & $0.271 \pm 0.065$ & $2.123 \pm 1.791$ & $\mathbf{0.776 \pm 0.125}$ & $0.571 \pm 0.018$ & $0.924 \pm 0.134$ & $\mathbf{0.669}$ \\
\bottomrule
    \end{tabular}
\end{table}

\begin{table}[h]
    \centering
    \scriptsize
    \caption{
        Acquisition performance on MIMIC Symile for AUPRC (Image imputed by ECG), showing $G_{\text{full}}$. Strategies are grouped by category. Best strategy among proposed and baseline methods in bold for each column.
    }
    \label{tab:acquisition_strategy_performance_auprc_mimic_haim_m0m2_revised_v2}
    \begin{tabular}{@{}l ccccc c@{}}
        \toprule
        \textbf{Strategy} & \textbf{Fracture} & \textbf{Enl. Card.} & \textbf{Consolidation} & \textbf{Atelectasis} & \textbf{Edema} & \textbf{Mean} \\
        \midrule
        \multicolumn{7}{@{}l}{\textit{Upper Bounds (for reference)}} \\
        Oracle & $6.343 \pm 5.165$ & $1.593 \pm 0.113$ & $1.950 \pm 0.162$ & $2.785 \pm 0.530$ & $1.832 \pm 0.041$ & $4.085$ \\
        True KL-Div. & $1.358 \pm 0.553$ & $0.818 \pm 0.060$ & $0.776 \pm 0.039$ & $0.526 \pm 0.206$ & $0.855 \pm 0.004$ & $0.814$ \\
        True Rank & $-0.157 \pm 0.576$ & $0.643 \pm 0.064$ & $0.808 \pm 0.008$ & $-0.477 \pm 0.824$ & $0.741 \pm 0.031$ & $0.420$ \\
        True Uncert. & $1.331 \pm 0.520$ & $0.699 \pm 0.124$ & $0.790 \pm 0.026$ & $-0.640 \pm 0.325$ & $0.745 \pm 0.040$ & $0.489$ \\
        \addlinespace
        \midrule 
        \multicolumn{7}{@{}l}{\textit{Imputation-based (proposed)}} \\
        KL-Div & $-0.784 \pm 1.571$ & $\mathbf{0.613 \pm 0.082}$ & $0.870 \pm 0.127$ & $-0.203 \pm 0.570$ & $\mathbf{0.737 \pm 0.035}$ & $0.434$ \\
\bottomrule
    \end{tabular}
    \vspace{1em} % Space between tabulars
    \begin{tabular}{@{}l ccccc c@{}}
        \toprule
        \textbf{Strategy} & \textbf{Cardiomegaly} & \textbf{Lung Lesion} & \textbf{Lung Opacity} & \textbf{Pneumonia} & \textbf{Pneumothorax} & \textbf{Mean} \\
        \midrule
        \multicolumn{7}{@{}l}{\textit{Upper Bounds (for reference)}} \\
        Oracle & $2.458 \pm 0.268$ & $2.158 \pm 0.482$ & $2.735 \pm 0.464$ & $3.162 \pm 0.383$ & $15.830 \pm 5.601$ & $4.085$ \\
        True KL-Div. & $0.827 \pm 0.009$ & $0.679 \pm 0.250$ & $0.711 \pm 0.063$ & $0.727 \pm 0.026$ & $0.861 \pm 0.024$ & $0.814$ \\
        True Rank & $0.555 \pm 0.041$ & $-0.005 \pm 0.389$ & $0.715 \pm 0.109$ & $0.809 \pm 0.034$ & $0.573 \pm 0.150$ & $0.420$ \\
        True Uncert. & $0.803 \pm 0.012$ & $0.184 \pm 0.270$ & $-0.069 \pm 0.111$ & $0.723 \pm 0.046$ & $0.326 \pm 0.207$ & $0.489$ \\
        \addlinespace
        \midrule 
        \multicolumn{7}{@{}l}{\textit{Imputation-based (proposed)}} \\
        KL-Div & $0.288 \pm 0.055$ & $0.468 \pm 0.419$ & $0.466 \pm 0.061$ & $0.595 \pm 0.049$ & $\mathbf{1.285 \pm 0.172}$ & $0.434$ \\
\bottomrule
    \end{tabular}
\end{table}

\begin{table}[h]
    \centering
    \scriptsize
    \caption{
        Acquisition performance on MIMIC Symile for AUROC (Image imputed by Lab and ECG), showing $G_{\text{full}}$. Strategies are grouped by category. Best strategy among proposed and baseline methods in bold for each column.
    }
    \label{tab:acquisition_strategy_performance_auroc_mimic_haim_m0m12_revised_v2}
    \begin{tabular}{@{}l ccccc c@{}}
        \toprule
        \textbf{Strategy} & \textbf{Fracture} & \textbf{Enl. Card.} & \textbf{Consolidation} & \textbf{Atelectasis} & \textbf{Edema} & \textbf{Mean} \\
        \midrule
        \multicolumn{7}{@{}l}{\textit{Upper Bounds (for reference)}} \\
        Oracle & $2.679 \pm 0.496$ & $3.081 \pm 0.257$ & $2.539 \pm 0.150$ & $5.098 \pm 0.624$ & $2.119 \pm 0.086$ & $3.911$ \\
        True KL-Div. & $1.006 \pm 0.111$ & $1.105 \pm 0.082$ & $0.880 \pm 0.033$ & $0.703 \pm 0.095$ & $1.017 \pm 0.023$ & $0.922$ \\
        True Rank & $0.760 \pm 0.142$ & $1.076 \pm 0.073$ & $0.993 \pm 0.052$ & $0.930 \pm 0.136$ & $1.036 \pm 0.023$ & $0.884$ \\
        True Uncert. & $0.975 \pm 0.140$ & $0.877 \pm 0.083$ & $0.886 \pm 0.065$ & $0.300 \pm 0.043$ & $0.634 \pm 0.018$ & $0.494$ \\
        \addlinespace
        \midrule 
        \multicolumn{7}{@{}l}{\textit{Imputation-based (proposed)}} \\
        KL-Div & $0.770 \pm 0.122$ & $\mathbf{0.736 \pm 0.077}$ & $0.711 \pm 0.031$ & $0.049 \pm 0.191$ & $0.535 \pm 0.015$ & $0.512$ \\
\bottomrule
    \end{tabular}
    \vspace{1em} % Space between tabulars
    \begin{tabular}{@{}l ccccc c@{}}
        \toprule
        \textbf{Strategy} & \textbf{Cardiomegaly} & \textbf{Lung Lesion} & \textbf{Lung Opacity} & \textbf{Pneumonia} & \textbf{Pneumothorax} & \textbf{Mean} \\
        \midrule
        \multicolumn{7}{@{}l}{\textit{Upper Bounds (for reference)}} \\
        Oracle & $2.552 \pm 0.116$ & $3.249 \pm 0.605$ & $4.964 \pm 0.431$ & $3.736 \pm 0.369$ & $9.096 \pm 1.223$ & $3.911$ \\
        True KL-Div. & $0.910 \pm 0.030$ & $0.517 \pm 0.243$ & $0.811 \pm 0.052$ & $0.967 \pm 0.061$ & $1.298 \pm 0.121$ & $0.922$ \\
        True Rank & $1.047 \pm 0.032$ & $0.545 \pm 0.093$ & $0.839 \pm 0.055$ & $1.010 \pm 0.056$ & $0.607 \pm 0.109$ & $0.884$ \\
        True Uncert. & $0.514 \pm 0.027$ & $0.200 \pm 0.151$ & $0.293 \pm 0.058$ & $0.729 \pm 0.044$ & $-0.468 \pm 0.155$ & $0.494$ \\
        \addlinespace
        \midrule 
        \multicolumn{7}{@{}l}{\textit{Imputation-based (proposed)}} \\
        KL-Div & $0.374 \pm 0.016$ & $0.403 \pm 0.210$ & $0.457 \pm 0.079$ & $0.518 \pm 0.036$ & $0.565 \pm 0.103$ & $0.512$ \\
\bottomrule
    \end{tabular}
\end{table}

\begin{table}[h]
    \centering
    \scriptsize
    \caption{
        Acquisition performance on MIMIC Symile for AUPRC (Image imputed by Lab and ECG), showing $G_{\text{full}}$. Strategies are grouped by category. Best strategy among proposed and baseline methods in bold for each column.
    }
    \label{tab:acquisition_strategy_performance_auprc_mimic_haim_m0m12_revised_v2}
    \begin{tabular}{@{}l ccccc c@{}}
        \toprule
        \textbf{Strategy} & \textbf{Fracture} & \textbf{Enl. Card.} & \textbf{Consolidation} & \textbf{Atelectasis} & \textbf{Edema} & \textbf{Mean} \\
        \midrule
        \multicolumn{7}{@{}l}{\textit{Upper Bounds (for reference)}} \\
        Oracle & $3.244 \pm 0.899$ & $2.660 \pm 0.365$ & $3.290 \pm 0.320$ & $3.756 \pm 0.538$ & $2.484 \pm 0.159$ & $4.063$ \\
        True KL-Div. & $1.111 \pm 0.223$ & $0.983 \pm 1.014$ & $0.998 \pm 0.079$ & $0.674 \pm 0.088$ & $1.014 \pm 0.034$ & $0.939$ \\
        True Rank & $0.503 \pm 0.538$ & $0.727 \pm 0.084$ & $1.063 \pm 0.108$ & $0.762 \pm 0.088$ & $1.014 \pm 0.036$ & $0.822$ \\
        True Uncert. & $0.826 \pm 0.166$ & $0.950 \pm 0.103$ & $0.923 \pm 0.060$ & $0.143 \pm 0.041$ & $0.636 \pm 0.029$ & $0.493$ \\
        \addlinespace
        \midrule 
        \multicolumn{7}{@{}l}{\textit{Imputation-based (proposed)}} \\
        KL-Div & $\mathbf{1.238 \pm 0.634}$ & $\mathbf{0.636 \pm 0.075}$ & $0.754 \pm 0.075$ & $0.330 \pm 0.146$ & $0.467 \pm 0.020$ & $0.585$ \\
\bottomrule
    \end{tabular}
    \vspace{1em} % Space between tabulars
    \begin{tabular}{@{}l ccccc c@{}}
        \toprule
        \textbf{Strategy} & \textbf{Cardiomegaly} & \textbf{Lung Lesion} & \textbf{Lung Opacity} & \textbf{Pneumonia} & \textbf{Pneumothorax} & \textbf{Mean} \\
        \midrule
        \multicolumn{7}{@{}l}{\textit{Upper Bounds (for reference)}} \\
        Oracle & $2.494 \pm 0.181$ & $2.376 \pm 0.472$ & $4.061 \pm 0.416$ & $4.964 \pm 0.520$ & $11.303 \pm 0.974$ & $4.063$ \\
        True KL-Div. & $0.823 \pm 0.022$ & $0.809 \pm 0.139$ & $0.873 \pm 0.081$ & $1.002 \pm 0.049$ & $1.105 \pm 0.058$ & $0.939$ \\
        True Rank & $1.012 \pm 0.039$ & $0.573 \pm 0.099$ & $0.952 \pm 0.093$ & $1.027 \pm 0.058$ & $0.591 \pm 0.077$ & $0.822$ \\
        True Uncert. & $0.394 \pm 0.037$ & $0.201 \pm 0.029$ & $0.184 \pm 0.050$ & $0.624 \pm 0.044$ & $0.054 \pm 0.058$ & $0.493$ \\
        \addlinespace
        \midrule 
        \multicolumn{7}{@{}l}{\textit{Imputation-based (proposed)}} \\
        KL-Div & $0.346 \pm 0.023$ & $0.560 \pm 0.075$ & $0.524 \pm 0.074$ & $0.441 \pm 0.037$ & $0.551 \pm 0.075$ & $0.585$ \\
\bottomrule
    \end{tabular}
\end{table}

\begin{table}[h]
    \centering
    \scriptsize
    \caption{
        Acquisition performance on MIMIC Symile for AUROC (Lab imputed by Image), showing $G_{\text{full}}$. Strategies are grouped by category. Best strategy among proposed and baseline methods in bold for each column.
    }
    \label{tab:acquisition_strategy_performance_auroc_mimic_haim_m1m0_revised_v2}
    \begin{tabular}{@{}l ccccc c@{}}
        \toprule
        \textbf{Strategy} & \textbf{Fracture} & \textbf{Enl. Card.} & \textbf{Consolidation} & \textbf{Atelectasis} & \textbf{Edema} & \textbf{Mean} \\
        \midrule
        \multicolumn{7}{@{}l}{\textit{Upper Bounds (for reference)}} \\
        Oracle & $2.404 \pm 0.448$ & $5.403 \pm 1.547$ & $2.446 \pm 0.350$ & $4.261$ & $4.715 \pm 0.988$ & $6.306$ \\
        True KL-Div. & $0.901 \pm 0.104$ & $0.727 \pm 0.113$ & $0.874 \pm 0.045$ & $0.807$ & $0.670 \pm 0.093$ & $0.634$ \\
        True Rank & $0.490 \pm 0.208$ & $0.872 \pm 0.019$ & $0.873 \pm 0.081$ & $1.691$ & $0.627 \pm 0.148$ & $0.604$ \\
        True Uncert. & $-0.204 \pm 0.122$ & $0.690 \pm 0.193$ & $0.319 \pm 0.084$ & $-0.511$ & $0.507 \pm 0.100$ & $0.154$ \\
        \addlinespace
        \midrule 
        \multicolumn{7}{@{}l}{\textit{Imputation-based (proposed)}} \\
        KL-Div & $0.443 \pm 0.028$ & $0.503 \pm 0.316$ & $0.523 \pm 0.078$ & $\mathbf{0.383}$ & $0.193 \pm 0.090$ & $\mathbf{0.684}$ \\
        Probability & $0.087 \pm 0.217$ & $0.428 \pm 0.155$ & $\mathbf{0.574 \pm 0.117}$ & $0.221$ & $0.457 \pm 0.056$ & $0.380$ \\
        Rank & $\mathbf{0.848 \pm 0.131}$ & $-0.103 \pm 0.100$ & $0.309 \pm 0.073$ & $-0.390$ & $0.463 \pm 0.125$ & $-0.139$ \\
        Uncertainty & $0.074 \pm 0.187$ & $0.593 \pm 0.168$ & $0.415 \pm 0.107$ & $0.278$ & $0.624 \pm 0.029$ & $0.311$ \\
        \addlinespace
        \midrule 
        \multicolumn{7}{@{}l}{\textit{Baselines (no imputation)}} \\
        Uncertainty & $0.515 \pm 0.062$ & $\mathbf{0.781 \pm 0.101}$ & $0.394 \pm 0.106$ & $0.310$ & $\mathbf{0.636 \pm 0.041}$ & $0.487$ \\
        Probability & $0.105 \pm 0.251$ & $0.444 \pm 0.144$ & $0.567 \pm 0.112$ & $0.267$ & $0.449 \pm 0.055$ & $0.390$ \\
        Random & $0.366 \pm 0.519$ & $0.662 \pm 0.085$ & $0.504 \pm 0.111$ & $0.087$ & $0.229 \pm 0.142$ & $0.419$ \\
        \bottomrule
    \end{tabular}
    \vspace{1em} % Space between tabulars
    \begin{tabular}{@{}l ccccc c@{}}
        \toprule
        \textbf{Strategy} & \textbf{Cardiomegaly} & \textbf{Lung Lesion} & \textbf{Lung Opacity} & \textbf{Pneumonia} & \textbf{Pneumothorax} & \textbf{Mean} \\
        \midrule
        \multicolumn{7}{@{}l}{\textit{Upper Bounds (for reference)}} \\
        Oracle & $11.461$ & $9.175 \pm 4.619$ & $10.440 \pm 4.341$ & $4.058 \pm 0.233$ & $8.693 \pm 2.295$ & $6.306$ \\
        True KL-Div. & $0.359$ & $0.043 \pm 1.152$ & $0.608 \pm 0.141$ & $0.793 \pm 0.054$ & $0.561 \pm 0.157$ & $0.634$ \\
        True Rank & $0.396$ & $0.640 \pm 1.426$ & $-0.637 \pm 0.091$ & $0.797 \pm 0.031$ & $0.296 \pm 0.208$ & $0.604$ \\
        True Uncert. & $0.507$ & $1.296 \pm 0.137$ & $-1.510 \pm 0.311$ & $0.320 \pm 0.119$ & $0.128 \pm 0.257$ & $0.154$ \\
        \addlinespace
        \midrule 
        \multicolumn{7}{@{}l}{\textit{Imputation-based (proposed)}} \\
        KL-Div & $0.477$ & $\mathbf{3.593 \pm 0.801}$ & $0.055 \pm 1.005$ & $\mathbf{0.630 \pm 0.024}$ & $0.037 \pm 0.291$ & $\mathbf{0.684}$ \\
        Probability & $0.551$ & $2.347 \pm 0.438$ & $-2.017 \pm 0.592$ & $0.581 \pm 0.030$ & $\mathbf{0.575 \pm 0.116}$ & $0.380$ \\
        Rank & $-0.413$ & $-1.673 \pm 4.164$ & $-0.741 \pm 0.142$ & $0.398 \pm 0.150$ & $-0.084 \pm 0.300$ & $-0.139$ \\
        Uncertainty & $\mathbf{0.646}$ & $0.200 \pm 0.174$ & $-0.684 \pm 0.697$ & $0.401 \pm 0.073$ & $0.566 \pm 0.142$ & $0.311$ \\
        \addlinespace
        \midrule 
        \multicolumn{7}{@{}l}{\textit{Baselines (no imputation)}} \\
        Uncertainty & $0.277$ & $0.172 \pm 0.139$ & $\mathbf{0.706 \pm 0.448}$ & $0.501 \pm 0.107$ & $0.575 \pm 0.126$ & $0.487$ \\
        Probability & $0.471$ & $1.854 \pm 1.407$ & $-1.445 \pm 0.962$ & $0.615 \pm 0.016$ & $0.575 \pm 0.126$ & $0.390$ \\
        Random & $-0.872$ & $2.030 \pm 1.179$ & $0.329 \pm 0.460$ & $0.490 \pm 0.124$ & $0.365 \pm 0.137$ & $0.419$ \\
        \bottomrule
    \end{tabular}
\end{table}

\begin{table}[h]
    \centering
    \scriptsize
    \caption{
        Acquisition performance on MIMIC Symile for AUPRC (Lab imputed by Image), showing $G_{\text{full}}$. Strategies are grouped by category. Best strategy among proposed and baseline methods in bold for each column.
    }
    \label{tab:acquisition_strategy_performance_auprc_mimic_haim_m1m0_revised_v2}
    \begin{tabular}{@{}l ccccc c@{}}
        \toprule
        \textbf{Strategy} & \textbf{Fracture} & \textbf{Enl. Card.} & \textbf{Consolidation} & \textbf{Atelectasis} & \textbf{Edema} & \textbf{Mean} \\
        \midrule
        \multicolumn{7}{@{}l}{\textit{Upper Bounds (for reference)}} \\
        Oracle & $2.142 \pm 0.360$ & $3.772 \pm 0.624$ & $4.042 \pm 1.747$ & -- & $5.560 \pm 1.093$ & $4.751$ \\
        True KL-Div. & $0.916 \pm 0.056$ & $0.648 \pm 0.079$ & $0.997 \pm 0.160$ & -- & $0.568 \pm 0.132$ & $0.699$ \\
        True Rank & $0.400 \pm 0.206$ & $0.686 \pm 0.095$ & $1.106 \pm 0.342$ & -- & $0.603 \pm 0.139$ & $0.590$ \\
        True Uncert. & $-1.106 \pm 0.446$ & $0.506 \pm 0.100$ & $0.258 \pm 0.136$ & -- & $0.477 \pm 0.270$ & $0.153$ \\
        \addlinespace
        \midrule 
        \multicolumn{7}{@{}l}{\textit{Imputation-based (proposed)}} \\
        KL-Div & $0.499 \pm 0.117$ & $0.481 \pm 0.203$ & $0.413 \pm 0.187$ & -- & $0.269 \pm 0.189$ & $\mathbf{0.680}$ \\
        Probability & $-0.381 \pm 0.441$ & $0.618 \pm 0.008$ & $0.339 \pm 0.318$ & -- & $\mathbf{0.681 \pm 0.114}$ & $0.442$ \\
        Rank & $\mathbf{0.824 \pm 0.103}$ & $-0.157 \pm 0.006$ & $-0.044 \pm 0.404$ & -- & $0.246 \pm 0.115$ & $0.186$ \\
        Uncertainty & $-0.429 \pm 0.473$ & $0.415 \pm 0.161$ & $\mathbf{0.541 \pm 0.220}$ & -- & $0.570 \pm 0.164$ & $0.124$ \\
        \addlinespace
        \midrule 
        \multicolumn{7}{@{}l}{\textit{Baselines (no imputation)}} \\
        Uncertainty & $0.455 \pm 0.060$ & $0.484 \pm 0.099$ & $0.537 \pm 0.231$ & -- & $0.585 \pm 0.160$ & $0.440$ \\
        Probability & $-0.362 \pm 0.445$ & $0.629 \pm 0.007$ & $0.325 \pm 0.326$ & -- & $0.672 \pm 0.112$ & $0.572$ \\
        Random & $0.283 \pm 0.513$ & $\mathbf{0.635 \pm 0.001}$ & $0.303 \pm 0.231$ & -- & $0.253 \pm 0.115$ & $0.356$ \\
        \bottomrule
    \end{tabular}
    \vspace{1em} % Space between tabulars
    \begin{tabular}{@{}l ccccc c@{}}
        \toprule
        \textbf{Strategy} & \textbf{Cardiomegaly} & \textbf{Lung Lesion} & \textbf{Lung Opacity} & \textbf{Pneumonia} & \textbf{Pneumothorax} & \textbf{Mean} \\
        \midrule
        \multicolumn{7}{@{}l}{\textit{Upper Bounds (for reference)}} \\
        Oracle & -- & $3.615 \pm 0.753$ & $3.868$ & $4.819 \pm 0.587$ & $10.192 \pm 4.607$ & $4.751$ \\
        True KL-Div. & -- & $0.503 \pm 0.458$ & $0.513$ & $0.865 \pm 0.112$ & $0.585 \pm 0.087$ & $0.699$ \\
        True Rank & -- & $0.974 \pm 0.661$ & $-0.184$ & $0.709 \pm 0.066$ & $0.426 \pm 0.076$ & $0.590$ \\
        True Uncert. & -- & $0.621 \pm 0.157$ & $-0.406$ & $0.202 \pm 0.069$ & $0.674 \pm 0.151$ & $0.153$ \\
        \addlinespace
        \midrule 
        \multicolumn{7}{@{}l}{\textit{Imputation-based (proposed)}} \\
        KL-Div & -- & $\mathbf{1.905 \pm 0.140}$ & $\mathbf{0.987}$ & $\mathbf{0.723 \pm 0.115}$ & $0.162 \pm 0.123$ & $\mathbf{0.680}$ \\
        Probability & -- & $1.133 \pm 0.483$ & $-0.486$ & $0.659 \pm 0.050$ & $\mathbf{0.976 \pm 0.085}$ & $0.442$ \\
        Rank & -- & $0.274 \pm 1.456$ & $-0.263$ & $0.304 \pm 0.238$ & $0.301 \pm 0.113$ & $0.186$ \\
        Uncertainty & -- & $-0.397 \pm 0.271$ & $-0.893$ & $0.261 \pm 0.024$ & $0.925 \pm 0.045$ & $0.124$ \\
        \addlinespace
        \midrule 
        \multicolumn{7}{@{}l}{\textit{Baselines (no imputation)}} \\
        Uncertainty & -- & $0.053 \pm 0.070$ & $0.109$ & $0.330 \pm 0.055$ & $0.963 \pm 0.071$ & $0.440$ \\
        Probability & -- & $1.363 \pm 0.660$ & $0.282$ & $0.708 \pm 0.061$ & $0.963 \pm 0.071$ & $0.572$ \\
        Random & -- & $0.088 \pm 0.319$ & $0.534$ & $0.455 \pm 0.187$ & $0.296 \pm 0.199$ & $0.356$ \\
        \bottomrule
    \end{tabular}
\end{table}

\begin{table}[h]
    \centering
    \scriptsize
    \caption{
        Acquisition performance on MIMIC Symile for AUROC (Lab imputed by ECG), showing $G_{\text{full}}$. Strategies are grouped by category. Best strategy among proposed and baseline methods in bold for each column.
    }
    \label{tab:acquisition_strategy_performance_auroc_mimic_haim_m1m2_revised_v2}
    \begin{tabular}{@{}l ccccc c@{}}
        \toprule
        \textbf{Strategy} & \textbf{Fracture} & \textbf{Enl. Card.} & \textbf{Consolidation} & \textbf{Atelectasis} & \textbf{Edema} & \textbf{Mean} \\
        \midrule
        \multicolumn{7}{@{}l}{\textit{Upper Bounds (for reference)}} \\
        Oracle & $2.830 \pm 1.006$ & $6.507 \pm 2.813$ & $1.993 \pm 0.252$ & $13.751 \pm 3.349$ & $5.706 \pm 1.964$ & $7.637$ \\
        True KL-Div. & $0.807 \pm 0.072$ & $1.043 \pm 0.176$ & $0.855 \pm 0.056$ & $0.933 \pm 0.345$ & $0.841 \pm 0.028$ & $0.727$ \\
        True Rank & $0.454 \pm 0.258$ & $0.880 \pm 0.255$ & $0.756 \pm 0.027$ & $0.462 \pm 0.641$ & $0.398 \pm 0.151$ & $0.290$ \\
        True Uncert. & $0.665 \pm 0.104$ & $0.273 \pm 0.305$ & $0.688 \pm 0.056$ & $-0.218 \pm 1.149$ & $0.568 \pm 0.041$ & $0.437$ \\
        \addlinespace
        \midrule 
        \multicolumn{7}{@{}l}{\textit{Imputation-based (proposed)}} \\
        KL-Div & $\mathbf{0.735 \pm 0.130}$ & $\mathbf{1.221 \pm 0.208}$ & $0.670 \pm 0.136$ & $\mathbf{4.739 \pm 1.683}$ & $2.234 \pm 0.684$ & $\mathbf{2.096}$ \\
\bottomrule
    \end{tabular}
    \vspace{1em} % Space between tabulars
    \begin{tabular}{@{}l ccccc c@{}}
        \toprule
        \textbf{Strategy} & \textbf{Cardiomegaly} & \textbf{Lung Lesion} & \textbf{Lung Opacity} & \textbf{Pneumonia} & \textbf{Pneumothorax} & \textbf{Mean} \\
        \midrule
        \multicolumn{7}{@{}l}{\textit{Upper Bounds (for reference)}} \\
        Oracle & -- & $13.135 \pm 7.362$ & $9.638 \pm 2.483$ & $3.075 \pm 0.250$ & $12.098 \pm 2.848$ & $7.637$ \\
        True KL-Div. & -- & $-0.560 \pm 1.556$ & $0.982 \pm 0.050$ & $0.761 \pm 0.031$ & $0.884 \pm 0.076$ & $0.727$ \\
        True Rank & -- & $-2.847 \pm 4.746$ & $1.545 \pm 0.340$ & $0.776 \pm 0.020$ & $0.185 \pm 0.511$ & $0.290$ \\
        True Uncert. & -- & $2.300 \pm 1.585$ & $0.157 \pm 0.299$ & $0.563 \pm 0.120$ & $-1.060 \pm 0.540$ & $0.437$ \\
        \addlinespace
        \midrule 
        \multicolumn{7}{@{}l}{\textit{Imputation-based (proposed)}} \\
        KL-Div & -- & $\mathbf{4.580 \pm 1.600}$ & $\mathbf{2.539 \pm 0.511}$ & $\mathbf{0.646 \pm 0.117}$ & $1.494 \pm 0.303$ & $\mathbf{2.096}$ \\
        Probability & -- & $-2.903 \pm 3.257$ & $-1.391 \pm 0.383$ & $0.455 \pm 0.124$ & $\mathbf{1.617 \pm 0.337}$ & $-0.333$ \\
        Rank & -- & $2.305 \pm 0.688$ & $1.567 \pm 0.771$ & $0.642 \pm 0.053$ & $0.017 \pm 0.983$ & $0.619$ \\
        Uncertainty & -- & $-2.336 \pm 1.344$ & $-1.290 \pm 0.341$ & $0.494 \pm 0.140$ & $-1.416 \pm 0.454$ & $-0.744$ \\
        \addlinespace
        \midrule 
        \multicolumn{7}{@{}l}{\textit{Baselines (no imputation)}} \\
        Uncertainty & -- & $1.397 \pm 1.699$ & $0.190 \pm 0.028$ & $0.541 \pm 0.082$ & $0.867 \pm 0.198$ & $0.574$ \\
        Probability & -- & $0.657 \pm 2.145$ & $1.107 \pm 0.654$ & $0.626 \pm 0.050$ & $0.867 \pm 0.198$ & $0.521$ \\
        Random & -- & $-0.793 \pm 3.159$ & $0.931 \pm 0.196$ & $0.555 \pm 0.085$ & $-0.086 \pm 0.456$ & $0.189$ \\
        \bottomrule
    \end{tabular}
\end{table}

\begin{table}[h]
    \centering
    \scriptsize
    \caption{
        Acquisition performance on MIMIC Symile for AUPRC (Lab imputed by ECG), showing $G_{\text{full}}$. Strategies are grouped by category. Best strategy among proposed and baseline methods in bold for each column.
    }
    \label{tab:acquisition_strategy_performance_auprc_mimic_haim_m1m2_revised_v2}
    \begin{tabular}{@{}l ccccc c@{}}
        \toprule
        \textbf{Strategy} & \textbf{Fracture} & \textbf{Enl. Card.} & \textbf{Consolidation} & \textbf{Atelectasis} & \textbf{Edema} & \textbf{Mean} \\
        \midrule
        \multicolumn{7}{@{}l}{\textit{Upper Bounds (for reference)}} \\
        Oracle & $4.231 \pm 2.318$ & $4.018 \pm 1.224$ & $2.093 \pm 0.267$ & $6.269 \pm 0.383$ & $5.741 \pm 1.374$ & $6.317$ \\
        True KL-Div. & $0.924 \pm 0.069$ & $0.825 \pm 0.017$ & $0.883 \pm 0.030$ & $0.717 \pm 0.570$ & $0.897 \pm 0.019$ & $0.972$ \\
        True Rank & $0.030 \pm 0.698$ & $0.643 \pm 0.103$ & $0.846 \pm 0.041$ & $0.619 \pm 1.260$ & $0.351 \pm 0.208$ & $0.900$ \\
        True Uncert. & $0.730 \pm 0.109$ & $0.793 \pm 0.209$ & $0.784 \pm 0.034$ & $0.030 \pm 0.615$ & $0.740 \pm 0.024$ & $0.550$ \\
        \addlinespace
        \midrule 
        \multicolumn{7}{@{}l}{\textit{Imputation-based (proposed)}} \\
        KL-Div & $\mathbf{0.721 \pm 0.166}$ & $\mathbf{1.292 \pm 0.411}$ & $0.780 \pm 0.157$ & $\mathbf{1.533 \pm 0.215}$ & $2.267 \pm 0.487$ & $\mathbf{1.779}$ \\
\bottomrule
    \end{tabular}%
\end{table}

\begin{table}[h]
    \centering
    \scriptsize
    \caption{
        Acquisition performance on MIMIC Symile for AUROC (Lab imputed by Image and ECG), showing $G_{\text{full}}$. Strategies are grouped by category. Best strategy among proposed and baseline methods in bold for each column.
    }
    \label{tab:acquisition_strategy_performance_auroc_mimic_haim_m1m02_revised_v2}
    \begin{tabular}{@{}l ccccc c@{}}
        \toprule
        \textbf{Strategy} & \textbf{Fracture} & \textbf{Enl. Card.} & \textbf{Consolidation} & \textbf{Atelectasis} & \textbf{Edema} & \textbf{Mean} \\
        \midrule
        \multicolumn{7}{@{}l}{\textit{Upper Bounds (for reference)}} \\
        Oracle & $2.416 \pm 0.727$ & $2.670 \pm 0.218$ & $3.090 \pm 0.355$ & $3.712 \pm 0.446$ & $2.174 \pm 0.050$ & $4.233$ \\
        True KL-Div. & $1.165 \pm 0.267$ & $0.883 \pm 0.024$ & $1.050 \pm 0.058$ & $1.092 \pm 0.080$ & $0.925 \pm 0.006$ & $0.981$ \\
        True Rank & $1.373 \pm 0.427$ & $0.811 \pm 0.024$ & $1.076 \pm 0.050$ & $0.904 \pm 0.066$ & $0.888 \pm 0.008$ & $0.883$ \\
        True Uncert. & $1.008 \pm 0.213$ & $0.720 \pm 0.056$ & $0.781 \pm 0.030$ & $-0.092 \pm 0.044$ & $0.675 \pm 0.027$ & $0.494$ \\
        \addlinespace
        \midrule 
        \multicolumn{7}{@{}l}{\textit{Imputation-based (proposed)}} \\
        KL-Div & $\mathbf{1.088 \pm 0.224}$ & $\mathbf{0.875 \pm 0.022}$ & $0.667 \pm 0.039$ & $\mathbf{1.193 \pm 0.091}$ & $\mathbf{0.885 \pm 0.008}$ & $\mathbf{1.035}$ \\
\bottomrule
    \end{tabular}%
\end{table}

\begin{table}[h]
    \centering
    \scriptsize
    \caption{
        Acquisition performance on MIMIC Symile for AUPRC (Lab imputed by Image and ECG), showing $G_{\text{full}}$. Strategies are grouped by category. Best strategy among proposed and baseline methods in bold for each column.
    }
    \label{tab:acquisition_strategy_performance_auprc_mimic_haim_m1m02_revised_v2}
    %
\end{table}

\begin{table}[h]
    \centering
    \scriptsize
    \caption{
        Acquisition performance on MIMIC Symile for AUROC (ECG imputed by Image), showing $G_{\text{full}}$. Strategies are grouped by category. Best strategy among proposed and baseline methods in bold for each column.
    }
    \label{tab:acquisition_strategy_performance_auroc_mimic_haim_m2m0_revised_v2}
    %
\end{table}

\begin{table}[h]
    \centering
    \scriptsize
    \caption{
        Acquisition performance on MIMIC Symile for AUPRC (ECG imputed by Image), showing $G_{\text{full}}$. Strategies are grouped by category. Best strategy among proposed and baseline methods in bold for each column.
    }
    \label{tab:acquisition_strategy_performance_auprc_mimic_haim_m2m0_revised_v2}
    %
\end{table}

\begin{table}[h]
    \centering
    \scriptsize
    \caption{
        Acquisition performance on MIMIC Symile for AUROC (ECG imputed by Lab), showing $G_{\text{full}}$. Strategies are grouped by category. Best strategy among proposed and baseline methods in bold for each column.
    }
    \label{tab:acquisition_strategy_performance_auroc_mimic_haim_m2m1_revised_v2}
    %
\end{table}

\begin{table}[h]
    \centering
    \scriptsize
    \caption{
        Acquisition performance on MIMIC Symile for AUPRC (ECG imputed by Lab), showing $G_{\text{full}}$. Strategies are grouped by category. Best strategy among proposed and baseline methods in bold for each column.
    }
    \label{tab:acquisition_strategy_performance_auprc_mimic_haim_m2m1_revised_v2}
    %
\end{table}

\begin{table}[h]
    \centering
    \scriptsize
    \caption{
        Acquisition performance on MIMIC Symile for AUROC (ECG imputed by Image and Lab), showing $G_{\text{full}}$. Strategies are grouped by category. Best strategy among proposed and baseline methods in bold for each column.
    }
    \label{tab:acquisition_strategy_performance_auroc_mimic_haim_m2m01_revised_v2}
    %
\end{table}

\begin{table}[h]
    \centering
    \scriptsize
    \caption{
        Acquisition performance on MIMIC Symile for AUPRC (ECG imputed by Image and Lab), showing $G_{\text{full}}$. Strategies are grouped by category. Best strategy among proposed and baseline methods in bold for each column.
    }
    \label{tab:acquisition_strategy_performance_auprc_mimic_haim_m2m01_revised_v2}
    %
\end{table}

\clearpage  
\section{Results for MIMIC HAIM}
\label{app:results_mimic_haim}
% Image, Lab
\begin{table}[h]
    \centering
    \scriptsize
    \caption{
        Acquisition performance on MIMIC HAIM for AUROC, showing $G_{\text{full}}$. Strategies are grouped by category. Best strategy among proposed ones and baselines in bold for each column.
    }
    \label{tab:acquisition_strategy_performance_auroc_mimic_haim_revised}
    %
\end{table}

\begin{table}[h]
    \centering
    \scriptsize
    \caption{
        Acquisition performance on MIMIC HAIM for AUPRC, showing $G_{\text{full}}$. Strategies are grouped by category. Best strategy among proposed ones and baselines in bold for each column.
    }
    \label{tab:acquisition_strategy_performance_auprc_mimic_haim_revised}
    %
\end{table}

\begin{table}[h]
    \centering
    \scriptsize
    \caption{
        Acquisition performance on MIMIC HAIM for AUROC (Image imputed by Lab), showing $G_{\text{full}}$. Strategies are grouped by category. Best strategy among proposed and baseline methods in bold for each column.
    }
    \label{tab:acquisition_strategy_performance_auroc_mimic_haim_m0m1_revised}
    %
\end{table}

\begin{table}[h]
    \centering
    \scriptsize
    \caption{
        Acquisition performance on MIMIC HAIM for AUPRC (Image imputed by Lab), showing $G_{\text{full}}$. Strategies are grouped by category. Best strategy among proposed and baseline methods in bold for each column.
    }
    \label{tab:acquisition_strategy_performance_auprc_mimic_haim_m0m1_revised}
    %
\end{table}

\begin{table}[h]
    \centering
    \scriptsize
    \caption{
        Acquisition performance on MIMIC HAIM for AUROC (Lab imputed by Image), showing $G_{\text{full}}$. Strategies are grouped by category. Best strategy among proposed and baseline methods in bold for each column.
    }
    \label{tab:acquisition_strategy_performance_auroc_mimic_haim_m1m0_revised}
    %
\end{table}

\begin{table}[h]
    \centering
    \scriptsize
    \caption{
        Acquisition performance on MIMIC HAIM for AUPRC (Lab imputed by Image), showing $G_{\text{full}}$. Strategies are grouped by category. Best strategy among proposed and baseline methods in bold for each column.
    }
    \label{tab:acquisition_strategy_performance_auprc_mimic_haim_m1m0_revised}
    %
\end{table}

%\newpage 
%\input{checklist.tex}

\end{document}